\documentclass{article}

\usepackage[final,nonatbib]{neurips_2020}

\usepackage{microtype}
\usepackage{graphicx}
\usepackage{booktabs}
\usepackage{hyperref}

\usepackage[utf8]{inputenc}
\usepackage[T1]{fontenc}
\usepackage{adjustbox}
\usepackage[noend]{algorithmic}
\usepackage{algorithmic}
\usepackage{amsfonts}
\usepackage[font=scriptsize]{subfig}
\usepackage{amsmath}
\usepackage{amssymb}
\usepackage{balance}
\usepackage{bbm}
\usepackage{bm}
\usepackage[font=small]{caption}
\usepackage{cite}
\usepackage{clipboard}
\usepackage{enumitem}
\usepackage{float}
\usepackage{makecell}
\usepackage{mathbbol}
\usepackage{mathtools, nccmath}
\usepackage{microtype}
\usepackage{multirow}
\usepackage{nicefrac}
\usepackage{pifont}
\usepackage{relsize}
\usepackage{setspace}
\usepackage{tcolorbox}
\usepackage{thmtools}
\usepackage{thm-restate}
\usepackage{url}
\usepackage{wrapfig}
\usepackage{xcolor}

\usepackage{algorithm}

\DeclareSymbolFontAlphabet{\mathbb}{AMSb}
\DeclareSymbolFontAlphabet{\mathbbl}{bbold}

\definecolor{crimson}{rgb}{0.7,0.01,0.02}
\definecolor{fern}{rgb}{0.05,0.6,0.05}
\definecolor{prussian}{rgb}{0,0.08,0.45}
\definecolor{faintgray}{gray}{0.9}
\definecolor{faintblue}{rgb}{0,0.08,0.65}

\newcommand{\pix}{\kern 0.1em}
\newcommand{\pmm}{\kern 0.25em$\pm$\kern 0.15em}
\newcommand{\pms}{\kern 0.10em$\pm$\kern 0.05em}
\newcommand{\cm}{\ding{51}}
\newcommand{\xm}{{\color{faintgray}\ding{55}}}
\newcommand{\na}{{\color{faintgray}(N/A)}}
\newcommand{\COM}[1]{\hfill$\triangleright$ #1}

\newcommand{\QED}{\hfill\raisebox{-0.5pt}{\scalebox{0.88}{$\square$}}}
\newcommand{\dayum}[1]{{#1\parfillskip=0pt\par}}

\newcommand{\smallsum}{\raisebox{1pt}{$\mathsmaller{\sum}$}}
\newcommand{\smallprod}{\raisebox{1pt}{$\mathsmaller{\prod}$}}

\declaretheorem[name=Theorem]{retheorem}
\declaretheorem[name=Proposition,numberlike=retheorem]{reproposition}
\declaretheorem[name=Lemma,numberlike=retheorem]{relemma}
\declaretheorem[name=Corollary,numberlike=retheorem]{recorollary}
\declaretheorem[name=Definition]{redefinition}

\let\OLDthebibliography\thebibliography

\renewcommand\thebibliography[1]{
  \OLDthebibliography{#1}
  \setlength{\itemsep}{5.8pt plus 2pt minus 1pt}
}

\usepackage{letltxmacro}

\newlength{\commentindent}
\makeatletter
\renewcommand{\algorithmiccomment}[1]{\unskip\hfill\makebox[\commentindent][l]{$\triangleright$~#1}\par}
\LetLtxMacro{\oldalgorithmic}{\algorithmic}
\renewcommand{\algorithmic}[1][0]{%
  \oldalgorithmic[#1]%
  \renewcommand{\ALC@com}[1]{%
    \ifnum\pdfstrcmp{##1}{default}=0\else\algorithmiccomment{##1}\fi}%
}
\makeatother

\newcommand{\asif}[2]{\rlap{$#1$}\hphantom{#2}}

\title{Time-series Generation by Contrastive Imitation}

\author{%
Daniel Jarrett\\
University of Cambridge, UK\\
\texttt{daniel.jarrett@maths.cam.ac.uk}
\And
Ioana Bica\\
University of Oxford, UK\\
Alan Turing Institute, UK\\
\texttt{ioana.bica@eng.ox.ac.uk}
\And
Mihaela van der Schaar\\
University of California, Los Angeles\\
University of Cambridge, UK\\
Alan Turing Institute, UK\\
\texttt{\small mv472@cam.ac.uk}
}

\begin{document}

\maketitle
\allowdisplaybreaks
\vspace{-0.50em}

\begin{abstract}

\dayum{
Consider learning a generative model for time-series data. The sequential setting poses a unique challenge: Not only should the generator capture the \textit{conditional} dynamics of (stepwise) transitions, but its open-loop rollouts should also preserve the \textit{joint} distribution of (multi-step) trajectories. On one hand, autoregressive models trained by MLE allow learning and computing explicit transition distributions, but suffer from compounding error during rollouts. On the other hand, adversarial models based on GAN training alleviate such exposure bias, but transitions are implicit and hard to assess. In this work, we study a generative framework that seeks to combine the strengths of both: Motivated by a moment-matching objective to mitigate compounding error, we optimize a local (but forward-looking) \textit{transition policy}, where the reinforcement signal is provided by a global (but stepwise-decomposable) \textit{energy model} trained by contrastive estimation. At training, the two components are learned cooperatively, avoiding the instabilities typical of adversarial objectives. At inference, the learned policy serves as the generator for iterative sampling, and the learned energy serves as a trajectory-level measure for evaluating sample quality. By expressly training a policy to imitate sequential behavior of time-series features in a dataset, this approach embodies ``\textit{generation by imitation}''. Theoretically, we illustrate the correctness of this formulation and the consistency of the algorithm. Empirically, we evaluate its ability to generate predictively useful samples from real-world datasets, verifying that it performs at the standard of existing benchmarks.}

\vspace{-0.50em}

\end{abstract}\section{Introduction}\label{sec:1}

\dayum{Time-series data are ubiquitous in diverse machine learning applications, such as financial, industrial, and healthcare settings. At the same time, lack of public access to data is a recurring obstacle to the development and reproducibility of research in domains where datasets are proprietary \cite{walonoski2018synthea}. Generating synthetic---but realistic---time-series data is a promising solution \cite{buczak2010data}, and has received increasing attention in recent years, driven by advances in deep learning and generative adversarial networks \cite{dash2020medical,jordon2021hide}.}

Owing to the fact that time-series features are generated sequentially, generative modeling in the temporal setting faces a two-pronged challenge: First, a good generator should accurately capture the conditional dynamics of \textit{stepwise} transitions $p(x_{t}|x_{1},...,x_{t-1})$; this is important, as the faithfulness of any conceivable downstream time-series analysis depends on the learned correlations across both temporal and feature dimensions. Second, however, the recursive rollouts of the generator should also respect the joint distribution of \textit{multi-step} trajectories $p(x_{1},...,x_{T})$; this is equally important, as synthetic trajectories that inadvertently wander beyond the support of original data are useless at best.

\dayum{Recent work falls into two main categories. On one hand, \textit{autoregressive models} trained via MLE \cite{williams1989learning} explicitly factor the distribution of trajectories into a product of conditionals $\smallprod_{t}p(x_{t}|x_{1},...,x_{t-1})$. While this allows directly learning and computing such transitions, with finite data this is prone to \textit{compounding errors} during multi-step generation, due to the discrepancy between closed-loop training (i.e. conditioned on ground-truths as inputs) and open-loop sampling (i.e. conditioned on its own previous outputs) \cite{ranzato2016sequence}. A variety of methods have sought to counteract this problem of exposure bias, employing auxiliary techniques from curriculum learning \cite{bengio1995input,bengio2009curriculum} and adversarial domain adaptation \cite{ganin2016domain,lamb2016professor}; however, such remedies are not without biases \cite{huszar2016not}, and empirical improvements have been mixed \cite{yoon2019time,goyal2017z,alaa2020generative}.}

\dayum{On the other hand, \textit{adversarial models} based on GAN training and its relatives \cite{gan, mirza2014conditional, genevay2018learning} directly model the distribution of trajectories $p(x_{1},...,x_{T})$ \cite{mogren2016c,lin2019generating,xu2020cot}. To provide a more granular learning signal for the generator, a popular variant matches the induced distribution of sub-trajectories instead, providing stepwise feedback from the discriminator \cite{esteban2017real,ramponi2018t}. TimeGAN \cite{yoon2019time} is the most recent incarnation of this, and operates within a jointly optimized latent space. GAN-based approaches alleviate the risk of compounding errors, and have been applied to banking \cite{tgan_bank}, sensors \cite{tgan_sensor}, biosignals \cite{tgan_biosignal}, and smartgrids \cite{tgan_smartgrid}. However, the conditional dynamics are only \textit{implicitly learned}, yielding no way of inspecting or assessing the quality of sampled transitions nor trajectories. Moreover, the adversarial objective leads to characteristically challenging optimization---exacerbated by the temporal dimension.}

\dayum{
\textbf{Three Operations}~
Consider a probabilistic generative model $p$ for some dataset $\mathcal{D}$. We are generally interested in performing one or more of the following operations: (1) \textit{sampling} a time series $\tau\sim p$, (2) \textit{evaluating} the likelihood $p(\tau)$, and (3) \textit{learning} the model $p$ from a set of i.i.d. samples $\tau$. In light of the preceding, we investigate a generative framework that attempts to fulfill the following criteria:}
\vspace{-0.5em}
\begin{itemize}[leftmargin=1.25em]
\itemsep-1pt
\item Samples should respect both the stepwise \textit{conditional} distributions of features, as well as the \textit{joint} distribution of full trajectories; unlike pure MLE, we wish to avoid multi-step compounding error.
\item Evaluating likelihoods should be possible as generic measures of \textit{sample quality} for both transitions and trajectories---often desired for sample comparison, model auditing, or bias correction \cite{alaa2021faithful,grover2019bias}.
\item \dayum{Unlike black-box GAN discriminators, we wish that the evaluator be \textit{decoupled} from any specific sampler, such that the two components can be trained \textit{non-adversarially},
thus may be more stable.}
\end{itemize}

\vspace{-0.50em}

\dayum{
\textbf{Contributions}~
In the sequel, we explore an approach that seeks to satisfy these criteria. We first give precise treatment of the ``compounding error'' problem, thus motivating a specific trajectory-centric optimization objective from first principles (Section \ref{sec:2}). To carry it out, we develop a general training framework and practical algorithm, along with its theoretical justification: We train a forward-looking \textit{transition policy} to imitate the sequential behavior of time series using a stepwise-decomposable \textit{energy model} as reinforcement, giving a method that embodies ``\textit{generation by imitation}'' (Section \ref{sec:3}). Importantly, to understand its strengths and limitations, we compare the method to existing generative models for time-series data, and relate it to imitation learning of sequential behavior (Section \ref{sec:4}). Lastly, through experiments with application to real-world time-series datasets, we verify that it generates predictively useful samples that perform at the standard of comparable benchmarks (Section \ref{sec:5}).}

\section{Synthetic Time Series}\label{sec:2}

\subsection{Problem Setup}\label{sec:21}

\dayum{
We operate in the standard discrete-time setting for time series. Let feature vectors $x_{t}\in\mathcal{X}$ be indexed by time steps $t$, and let a full trajectory of length $T$ be denoted $\tau\coloneqq(x_{1},...,x_{T})\in\mathcal{T}\coloneqq\mathcal{X}^{T}$. Also, denote with $h_{t}\coloneqq(x_{1},...,x_{t-1})\in\mathcal{H}\coloneqq\cup_{t=1}^{T}\mathcal{X}^{t}$ the history prior to time $t$. For ease of exposition we shall work with trajectories of fixed lengths $T$, but our results trivially generalize to the case where $T$ itself is a random variable (for instance, by employing padding tokens up to some maximum length).}

Consider a dataset $\mathcal{D}\coloneqq\{\tau_{n}\}_{n=1}^{N}$ of $N$ trajectories sampled from some true source $s$. We assume the trajectories are generated sequentially by some unknown transition process $\pi_{s}\in\Delta(\mathcal{X})^{\mathcal{H}}$, such that features at each step $t$ are sampled as $x_{t}\sim\pi_{s}(\cdot|h_{t})$. In addition to this stepwise conditional, denote with \smash{$\mu_{s}(h)\coloneqq\tfrac{1}{T}\sum_{t}p(h_{t}=h|\pi_{s})$} the normalized occupancy measure---i.e. the distribution of histories induced by $\pi_{s}$. Intuitively, this is the visitation distribution of ``history states'' encountered by a generator when navigating about the feature space $\mathcal{X}$ by rolling out policy $\pi_{s}$. With slight abuse of notation, we may also write $\mu_{s}(h,x)\coloneqq\mu_{s}(h)\pi_{s}(x|h)$ to indicate the marginal distribution of transitions. Finally, let the joint distribution of full trajectories be denoted by \smash{$p_{s}(\tau)\coloneqq\smallprod_{t}\pi_{s}(x_{t}|h_{t})$}.

The goal is to learn a sequential generator $\pi_{\theta}$ parameterized as $\theta$ using samples $\tau\sim p_{s}$ from $\mathcal{D}$, such that $p_{\theta}\approx p_{s}$. Note here that we do not assume stationarity of the time-series data, nor stationarity of the transition conditionals; any influence of $t$ is implicit through the dependence of $\pi_{s}$ (and $\pi_{\theta}$) on variable-length histories. In line with recent work \cite{alaa2020generative, xu2020cot}, for simplicity we do not consider static metadata as supplemental inputs or outputs, as these are commonly and easily incorporated via an additional conditioning layer or auxiliary generator \cite{yoon2019time, lin2019generating}. Lastly, note that much recent work on sequential modeling is devoted to domain-specific, \textit{architecture}-level designs for generating audio \cite{donahue2019adversarial,engel2019gansynth}, text \cite{nie2019relgan, caccia2020language}, and video \cite{saito2017temporal,tulyakov2018mocogan}. In contrast, our work is closer in spirit to \cite{yoon2019time,alaa2020generative} in being an agnostic, \textit{framework}-level study applicable to generic tabular data in any time-series setting.

\dayum{
\textbf{Measuring Sample Quality}~
How do we determine the ``quality'' of a sample? In specialized domains, of course, we often have prior access to \textit{task-specific} metrics such as \textsc{bleu} or \textsc{rouge} scores in text generation \cite{ranzato2016sequence,bahdanau2017actor}---then, the generator can simply be optimized for such scores via standard methods in reinforcement learning \cite{konda2000actor}. In generic time-series settings, however, the challenge is that any such metric must necessarily be \textit{task-agnostic}, and access to it must necessarily come from learning.}

\dayum{
So, for any data source $s$, let us speak of some hypothetical function $f_{s}:\mathcal{H}\times\mathcal{X}$\pix$\rightarrow$\pix$[-c,c]$ with $c<\infty$, such that $f_{s}(h,x)$ gives the quality of any sampled \textit{transition}---that is, any tuple $(h,x)$. Intuitively, we may interpret this as quantifying how ``typical'' it is for the random process to be in state $h$ and step towards $x$.
Likewise, let as also speak of some function $F_{s}:\mathcal{T}$\pix$\rightarrow$\pix$[-cT,cT]$ such that $F_{s}(\tau)$ gives the quality of any sampled \textit{trajectory}. Naturally, in time-series settings where the underlying process is causally-conditioned, it is reasonable to define this as the decomposition \smash{$F_{s}(\tau)\coloneqq\sum_{t}f_{s}(h_{t},x_{t})$}.
Now of course, we have no access to the true $F_{s}$. But clearly, in learning a generative model $p_{\theta}$ of $p_{s}$, we wish that the quality of samples $\tau$ drawn from $p_{\theta}$ and $p_{s}$ be similar in expectation. More precisely:}

\vspace{-0.25em}
\begin{redefinition}[restate=defdifference,name=Expected Quality Difference]\upshape\label{def:difference}
\dayum{
Let $\Delta\bar{F}_{s}$\pix$:$\pix$\Theta$\pix$\rightarrow$\pix$[-2cT,2cT]$ denote the \textit{expected quality difference} between $p_{s}$ and $p_{\theta}$, where $\Theta$ indicates the space of parameterizations for generator $\pi_{\theta}$:}
\vspace{-0.20em}
\begin{equation}
\Delta\bar{F}_{s}(\theta)
\coloneqq
\mathbb{E}_{\tau\sim p_{s}}F_{s}(\tau)
-
\mathbb{E}_{\tau\sim p_{\theta}}F_{s}(\tau)
\end{equation}
\end{redefinition}
\vspace{-0.45em}
\dayum{%
\hspace{-0.5pt}Our objective, then, is to learn a generator $\pi_{\theta}$ that minimizes the expected quality difference $\Delta\bar{F}_{s}(\theta)$.
Two points bear emphasis. First, we know nothing about $F_{s}$---beyond it being the sequential aggregate of $f_{s}$. This challenge uniquely differentiates this agnostic setting from more popular media-specific applications---for which various predefined measures are readily available for supervision. Second, in addition to matching this \textit{expectation} over samples, we also wish to match the \textit{variety} of samples in the original data. After all, we want $p_{\theta}$ to mimic samples from $p_{s}$ of different degrees of ``typicality''. So we should expect to incorporate some measure of entropy, e.g. the commonly used Shannon entropy.}

\subsection{Matching Local Moments}\label{sec:22}

\dayum{
Recall the apparent tradeoff between autoregressive models and adversarial models. In the spirit of the former, suppose we seek to directly learn \textit{transition conditionals} via supervised learning. That is,}

\vspace{-0.35em}

\begin{equation}\label{eq:local}
\text{arg\pix min}_{\theta}~
\mathbb{E}_{h\sim\mu_{s}}
\mathcal{L}(
\pi_{s}(\cdot|h),\pi_{\theta}(\cdot|h)
)
\end{equation}

\vspace{-0.15em}

\dayum{%
Consider the log likelihood loss \smash{$\mathcal{L}(\pi_{s}(\cdot|h),\pi_{\theta}(\cdot|h))\coloneqq-\mathbb{E}_{x\sim\pi_{s}(\cdot|h)}\log\pi_{\theta}(x|h)$}. In the case of exponential family models for $\pi_{\theta}(\cdot|h)$, a basic result is that this is dual to maximizing its conditional entropy subject to the constraint on feature expectations \smash{$\mathbb{E}_{h\sim\mu_{s};x\sim\pi_{\theta}(\cdot|h)}T(x)=\mathbb{E}_{h\sim\mu_{s};x\sim\pi_{s}(\cdot|h)}T(x)$}, where $T:\mathcal{X}\rightarrow\mathbb{R}$ is some sufficient statistic \cite{grunwald2004game,farnia2016minimax,ziebart2010thesis}. More generally for deep energy-based models, we have (however, recall that strong duality does not generalize to the nonlinear case; see Appendix \ref{app:a}):}

\vspace{-0.50em}

\begin{equation}\label{eq:local2}
\text{arg\pix min}_{\theta}~
\Big(
\mathbb{E}_{\substack{h\sim\mu_{s}\\x\sim \pi_{\theta}(\cdot|h)}}
\log\pi_{\theta}(x|h)
+
\text{max}_{f\in\mathbb{R}^{\mathcal{H}\times\mathcal{X}}}
\big(
\mathbb{E}_{\substack{h\sim\mu_{s}\\x\sim\pi_{s}(\cdot|h)}}
f(h,x)
-
\mathbb{E}_{\substack{h\sim\mu_{s}\\x\sim\pi_{\theta}(\cdot|h)}}
f(h,x)
\big)\Big)
\end{equation}
\dayum{%
Note that the moment-matching constraint is \textit{local}---that is, at the level of individual transitions, and all conditioning is based on $h$ from $\mu_{s}$ alone. This is precisely the ``exposure bias'': The objective is only ever exposed to inputs $h$ drawn from the (perfect) source distribution $\mu_{s}$, and is thus unaware of the endogeneity of the (imperfect) synthetic distribution $\mu_{\theta}$ induced by $\pi_{\theta}$. This is not desirable since $\pi_{\theta}$ is rolled out by open-loop sampling at test time. Now, although at the global optimum the moment-matching discrepancy must be zero (i.e. the equality constraint is enforced), in practice there may be a variety of reasons why this is not perfectly achieved (e.g. error in estimating expectations, error in function approximation, error in optimization, etc). Suppose we could bound how well we are able to enforce the moment-matching constraint; as it turns out, we cannot eliminate error compounding:\footnote{Lemmas \ref{thm:local} and \ref{thm:global} are similar in spirit to results for error accumulation in imitation by behavioral cloning and distribution matching.\pix~See Appendix \ref{app:a};\pix~this analogy with imitation learning is formally identified in Section \ref{sec:4}.}}

\begin{relemma}[restate=thmlocal,name=]\upshape\label{thm:local}
Let $\text{max}_{f\in\mathbb{R}^{\mathcal{H}\times\mathcal{X}}}
\big(
\mathbb{E}_{\substack{h\sim\mu_{s}\\x\sim\pi_{s}(\cdot|h)}}
f(h,x)
-
\mathbb{E}_{\substack{h\sim\mu_{s}\\x\sim\pi_{\theta}(\cdot|h)}}
f(h,x)
\big)
\leq
\epsilon$.
Then $\Delta\bar{F}_{s}(\theta)\in O(T^{2}\epsilon)$.
\end{relemma}

\vspace{-1.00em}

\textit{Proof}. Appendix \ref{app:a}. \QED

\vspace{0.25em}

This reveals the problem with modeling conditionals per se: \textit{Not all mistakes are equal}. An objective like Equation \ref{eq:local} penalizes unrealistic transitions $(h,x)$ by treating all conditioning histories $h$ equally---regardless of how realistic $h$ is to begin with. Clearly, however, we care much less about how $x$ looks like, if the current subsequence $h$ is already highly unlikely (and vice versa). Intuitively, earlier mistakes in a trajectory should weigh more: Once $\pi_{\theta}$ wanders into areas of $\mathcal{H}$ with low support in $\mu_{s}$, no amount of ``good'' transitions will bring the trajectory back to high-likelihood areas of $\mathcal{T}$ under $p_{s}$.

\subsection{Matching Global Moments}

\dayum{
Now suppose instead that we seek to directly constrain the \textit{trajectory distribution} $p_{\theta}$ to be similar to $p_{s}$:}

\vspace{-0.35em}

\begin{equation}\label{eq:global}
\text{arg\pix min}_{\theta}~
\mathcal{L}(
p_{s},p_{\theta}
)
\end{equation}

\vspace{-0.35em}

\dayum{%
Consider the Kullback-Leibler divergence $\mathcal{L}(p_{s},p_{\theta})\coloneqq D_{\text{KL}}(p_{s}\|p_{\theta})$. Like before, we know that in the case of exponential family models for $p_{\theta}$, this is dual to maximizing its entropy subject to the constraint $\mathbb{E}_{\tau\sim p_{s}}T(\tau)=\mathbb{E}_{\tau\sim p_{\theta}}T(\tau)$, where $T:\mathcal{T}\rightarrow\mathbb{R}$ is some sufficient statistic \cite{jordan2003introduction}. More broadly for deep energy-based models, we have $
\text{arg\pix min}_{\theta}~
(
\mathbb{E}_{\tau\sim p_{\theta}}
\log p_{\theta}(\tau)
+
\text{max}_{F\in\mathbb{R}^{\mathcal{T}}}~
(
\mathbb{E}_{\tau\sim p_{s}}
F(\tau)
-
\mathbb{E}_{\tau\sim p_{\theta}}
F(\tau)
))
$
(but again, recall here that strong duality does not generalize to the nonlinear case; see Appendix \ref{app:a}).
Now, observe that by definition of occupancy measure $\mu$, for any function $f:\mathcal{H}\times\mathcal{X}\rightarrow\mathbb{R}$ it must be the case that $
\mathbb{E}_{\tau\sim p}\sum_{t}f(h_{t},x_{t})
=
T\mathbb{E}_{h\sim\mu,x\sim\pi(\cdot|h)}
f(h,x)
$. Therefore we may equivalently write}

\vspace{-0.50em}

\begin{equation}\label{eq:global2}
\text{arg\pix min}_{\theta}~
\Big(
\mathbb{E}_{\substack{h\sim\mu_{\theta}\\x\sim\pi_{\theta}(\cdot|h)}}
\log\pi_{\theta}(x|h)
+
\text{max}_{f\in\mathbb{R}^{\mathcal{H}\times\mathcal{X}}}
\big(
\mathbb{E}_{\substack{h\sim\mu_{s}\\x\sim\pi_{s}(\cdot|h)}}
f(h,x)
-
\mathbb{E}_{\substack{h\sim\mu_{\theta}\\x\sim\pi_{\theta}(\cdot|h)}}
f(h,x)
\big)\Big)
\end{equation}
\dayum{%
Importantly, note that the moment-matching constraint is now \textit{global}---that is, at the level of trajectory rollouts, and $\pi_{\theta}$ is now conditioned on histories $h$ drawn from its own induced occupancy measure $\mu_{\theta}$. There is no longer any ``exposure bias'' here: In order to respect the constraint, not only does $\pi_{\theta}(\cdot|h)$ have to be close to $\pi_{s}(\cdot|h)$ for any given $h$, but the occupancy measure $\mu_{\theta}$ induced by $\pi_{\theta}$ also has to be close to the occupancy measure $\mu_{s}$ induced by $\pi_{s}$. As it turns out, this seemingly minor difference is sufficient to mitigate compounding errors. As before, although at the global optimum the moment-matching discrepancy must be zero, in practice this may not be perfectly achieved. Now, suppose we could bound how well we are able to enforce the moment-matching constraint; but we now have:}

\begin{relemma}[restate=thmglobal,name=]\upshape\label{thm:global}
\dayum{
Let $\text{max}_{f\in\mathbb{R}^{\mathcal{H}\times\mathcal{X}}}
\big(
\mathbb{E}_{\substack{h\sim\mu_{s}\\x\sim\pi_{s}(\cdot|h)}}
f(h,x)
-
\mathbb{E}_{\substack{h\sim\mu_{\theta}\\x\sim\pi_{\theta}(\cdot|h)}}
f(h,x)
\big)
\leq
\epsilon$.
Then $\Delta\bar{F}_{s}(\theta)\in O(T\epsilon)$.}
\end{relemma}

\vspace{-1.00em}

\textit{Proof}. Appendix \ref{app:a}. \QED

\vspace{0.25em}

\dayum{
This illustrates why even \textit{transition-centric} adversarial models such as \cite{yoon2019time,esteban2017real} have shown promise in generating realistic trajectories \cite{tgan_bank,tgan_sensor,tgan_biosignal,tgan_smartgrid}. First, unlike \textit{trajectory-centric} GANs \cite{mogren2016c,lin2019generating} which directly attempt to minimize some form of Equation \ref{eq:global}, in transition-centric GANs the objective is to match the transition marginals $\mu_{\theta}(h,x)$ and $\mu_{s}(h,x)$---so the discriminator provides more granular feedback to the generator for training. At the same time, we see from Lemma \ref{thm:global} that matching transition marginals is already---indirectly---performing the sort of moment-matching that alleviates compounding error.}

Can we be more direct? In Section \ref{sec:3}, we shall start by tackling Equation \ref{eq:global2} itself. As we shall see, this endeavor gives rise to a technique that trains a conditional policy (for sampling), an energy model (for evaluation), and a non-adversarial framework (for learning)---addressing our three initial criteria.

\section{Generating by Imitating}\label{sec:3}

\dayum{
First, consider the most straightforward implementation: Let us parameterize $f\in\mathbb{R}^{\mathcal{H}\times\mathcal{X}}$ as $\phi$, and begin with the primal form of Equation \ref{eq:global2}, which yields the following adversarial learning objective:}

\vspace{-0.25em}

\begin{equation}\label{eq:vebm}
\mathcal{L}(\theta,\phi)
\coloneqq
\text{max}_{\phi}~
\text{min}_{\theta}~
\Big(
\mathbb{E}_{\substack{h\sim\mu_{\theta}\\x\sim\pi_{\theta}(\cdot|h)}}
\log \pi_{\theta}(x|h)
+
\mathbb{E}_{\substack{h\sim\mu_{s}\\x\sim\pi_{s}(\cdot|h)}}
f_{\phi}(h,x)
-
\mathbb{E}_{\substack{h\sim\mu_{\theta}\\x\sim\pi_{\theta}(\cdot|h)}}
f_{\phi}(h,x)
\Big)
\end{equation}

\vspace{-0.50em}

\dayum{
It is easy to see that this effectively describes variational training of the energy-based model $p_{\phi}(\tau)\coloneqq$ $\exp(F_{\phi}(\tau)-\log Z_{\phi})$---where $F_{\phi}(\tau)\coloneqq\sum_{t}f_{\phi}(h_{t},x_{t})$---to approximate the true $p_{s}(\tau)$, using samples from the variational $p_{\theta}$. The (outer) energy player is the maximizing agent, and the (inner) policy player is the minimizing agent. The form of this objective naturally prescribes a bilevel optimization procedure in which we perform (gradient-based) updates of $\phi$ with nested (best-response) updates of $\theta$.}

\subsection{Challenges of Learning}\label{sec:31}

\dayum{
Abstractly, of course, training energy models using variational samplers is not new: Multiple works in static domains---such as image modeling---have investigated this approach as a means of bypassing the expense and variance of MCMC sampling \cite{kim2016deep,zhai2016generative}. In our setting, however, there is the additional \textit{temporal} dimension: The negative energy $F_{\phi}(\tau)$ of any trajectory is computed as the sequential composition of stepwise qualities $f_{\phi}(h_{t},x_{t})$, and each trajectory sampled from $p_{\theta}$ must be generated as the sequential rollout of stepwise policies $\pi_{\theta}(x_{t}|h_{t})$. Consider the gradient update for the energy,}

\vspace{-0.25em}

\begin{equation}\label{eq:vebmenergy}
\nabla_{\phi}\mathcal{L}
=
\mathbb{E}_{\substack{h\sim\mu_{s}\\x\sim\pi_{s}(\cdot|h)}}
\nabla_{\phi}
f_{\phi}(h,x)
-
\mathbb{E}_{\substack{h\sim\mu_{\theta}\\x\sim\pi_{\theta}(\cdot|h)}}
\nabla_{\phi}
f_{\phi}(h,x)
\end{equation}

\vspace{-0.50em}

and the inner-loop update for the policy,

\vspace{-0.50em}

\begin{equation}
\text{arg\pix min}_{\theta}~
\mathbb{E}_{\substack{h\sim\mu_{\theta}\\x\sim\pi_{\theta}(\cdot|h)}}
\log\pi_{\theta}(x|h)
-
\mathbb{E}_{\substack{h\sim\mu_{\theta}\\x\sim\pi_{\theta}(\cdot|h)}}
f_{\phi}(h,x)
\end{equation}

\vspace{-0.50em}

Note that the max-min optimization requires complete optimization within each inner update in order for the outer update to be correct. Otherwise the gradients will be \textit{biased}, and there would be no guarantee the procedure converges to anything meaningful. Yet unlike in the static setting---for which there exists variety of standard approximations for the inner update \cite{kim2016deep,zhai2016generative,dai2017calibrating,kumar2019maximum}---here the policy update amounts to entropy-regularized \textit{reinforcement learning} \cite{fox2016taming,haarnoja2017reinforcement,shi2019soft} using $f_{\phi}(h_{t},x_{t})$ as reward function. Thus our first difficulty is computational: Repeatedly performing inner-loop RL is simply infeasible.

\dayum{
Now, an obvious alternative is to dispense with complete policy optimization at each step, and instead to employ \textit{importance sampling} to ensure that the gradients for the energy updates are still unbiased:}

\vspace{-0.25em}

\begin{equation}
\nabla_{\phi}\mathcal{L}
=
\mathbb{E}_{\tau\sim p_{s}}
\nabla_{\phi}
F_{\phi}(\tau)
-
\mfrac{1}{Z_{\phi}}
\mathbb{E}_{\tau\sim p_{\theta}}
\Big[
\mfrac{\exp(\sum_{t}f_{\phi}(h_{t},x_{t}))}{\prod_{t}\pi_{\theta}(x_{t}|h_{t})}
\nabla_{\phi}
F_{\phi}(\tau)
\Big]
\end{equation}

\vspace{-0.35em}

\dayum{
where the partition function is computed as $
Z_{\phi}
=
\mathbb{E}_{\tau\sim p_{\theta}}
[\exp(\smallsum_{t}f_{\phi}(h_{t},x_{t}))/\smallprod_{t}\pi_{\theta}(x_{t}|h_{t})]
$, and the sampling policy $\pi_{\theta}$ is no longer required to be perfectly optimized with respect to $f_{\phi}$. Unfortunately, this strategy simply replaces the original difficulty with a statistical one: As soon as we consider time-series data of non-trivial lengths $T$, the multiplicative effect of each time step on the importance weights means the gradient estimates---albeit unbiased---will have impractically high variance \cite{sutton2018reinforcement,hanna2018towards}.}


\subsection{Contrastive Imitation}\label{sec:32}

\dayum{
We now investigate a generative framework that seeks to avoid these difficulties. The key idea is that instead of Equation \ref{eq:vebmenergy}, we shall learn $p_{\phi}$ by contrasting (real) ``positive'' samples $\tau\sim p_{s}$ and (any) ``negative'' samples $\tau\sim p_{\theta}$, which---as we shall see---rids us of the requirement that $\pi_{\theta}$ be fully optimized at each step for learning to be guaranteed. First, let us establish the notion of a ``structured classifier'':\footnote{The idea that density estimation can be performed by logistic regression goes back at least to \cite{hastie09unsupervised}, and formalized as negative sampling \cite{mikolov2013distributed} and noise-contrastive estimation \cite{gutmann2012noise}. Structured classifiers have been studied in the context of imitation learning \cite{finn2016connection,fu2018learning} by analogy with GANs. In the time-series setting, however, we shall see that this approach is equivalent to noise-contrastive estimation with an adaptive noise distribution.}}

\vspace{-0.50em}

\begin{redefinition}[restate=defclassifier,name=Structured Classification]\upshape\label{def:classifier}
\dayum{
Recall the $\pi_{\theta}$-induced distribution \smash{$p_{\theta}(\tau)\coloneqq\smallprod_{t}\pi_{\theta}(x_{t}|h_{t})$}. Denote with $\tilde{p}_{\phi}$ the \textit{un-normalized} energy-based model such that \smash{$\tilde{p}_{\phi}(\tau)\coloneqq\exp(\sum_{t}f_{\phi}(h_{t},x_{t}))$}, and let $Z_{\phi}$ be folded into $\phi$ as a learnable parameter. Define the \textit{structured classifier} $d_{\theta,\phi}:\mathcal{T}\rightarrow[0,1]$:}

\vspace{-0.65em}

\begin{equation}\label{eq:classifier}
d_{\theta,\phi}(\tau)
\coloneqq
\frac{\tfrac{1}{Z_{\phi}}\tilde{p}_{\phi}(\tau)}{\tfrac{1}{Z_{\phi}}\tilde{p}_{\phi}(\tau)+p_{\theta}(\tau)}
\end{equation}

\vspace{0.00em}

\end{redefinition}

\dayum{
That is, unlike a black-box classifier that may be arbitrarily parameterized---such as a generic discriminator $d$ in a GAN---here $d_{\theta,\phi}$ is ``structured'' in that it is modularly parameterized by the embedded energy and policy functions. Now, we shall train $\phi$ such that $d_{\theta,\phi}$ discriminates well between $\tau\sim p_{s}$ and $\tau\sim p_{\theta}$---that is, so that the output $d_{\theta,\phi}(\tau)$ represents the (posterior) probability that $\tau$ is real,}

\vspace{-0.05em}

\begin{equation}\label{eq:nceenergy}
\mathcal{L}_{\text{energy}}(\phi;\theta)
\coloneqq
\text{$-$}
\mathbb{E}_{\tau\sim p_{s}}
\log d_{\theta,\phi}(\tau)
\text{$-$}
\mathbb{E}_{\tau\sim p_{\theta}}
\log\big(1
\text{$-$}
d_{\theta,\phi}(\tau)\big)
\end{equation}

\vspace{-0.75em}

and as before,

\vspace{-1.00em}

\begin{equation}\label{eq:ncepolicy}
\mathcal{L}_{\text{policy}}(\theta;\phi)
\pix\coloneqq~~~
\mathbb{E}\hspace{-1px}_{\substack{h\sim\mu_{\theta}\\x\sim\pi_{\theta}(\cdot|h)}}\hspace{-2px}
\log\pi_{\theta}(x|h)
-
\mathbb{E}\hspace{-1px}_{\substack{h\sim\mu_{\theta}\\x\sim\pi_{\theta}(\cdot|h)}}\hspace{-1px}
f_{\phi}(h,x)
\end{equation}


\vspace{-0.50em}

\dayum{
Why is this better? As we now show formally, each gradient update no longer requires $\theta$ to be optimal for the current value of $\phi$---nor does it require importance sampling---unlike the procedure described by Equation \ref{eq:vebm}. The only requirement is that $p_{\theta}$ can be sampled and evaluated efficiently, e.g. using learned Gaussian policies as usual, or---should more flexibility be required---with normalizing flow-based policies. As a practical result, this means policy updates can be \textit{interleaved} with energy updates, instead of being \textit{nested} within a repeated inner loop. Specifically, let us establish the following results:}

\begin{figure}[t]
\vspace{-1.0em}
\centering
\makebox[\textwidth][c]{
\subfloat[T-Forcing \cite{williams1989learning}]
{\includegraphics[width=0.26\linewidth]{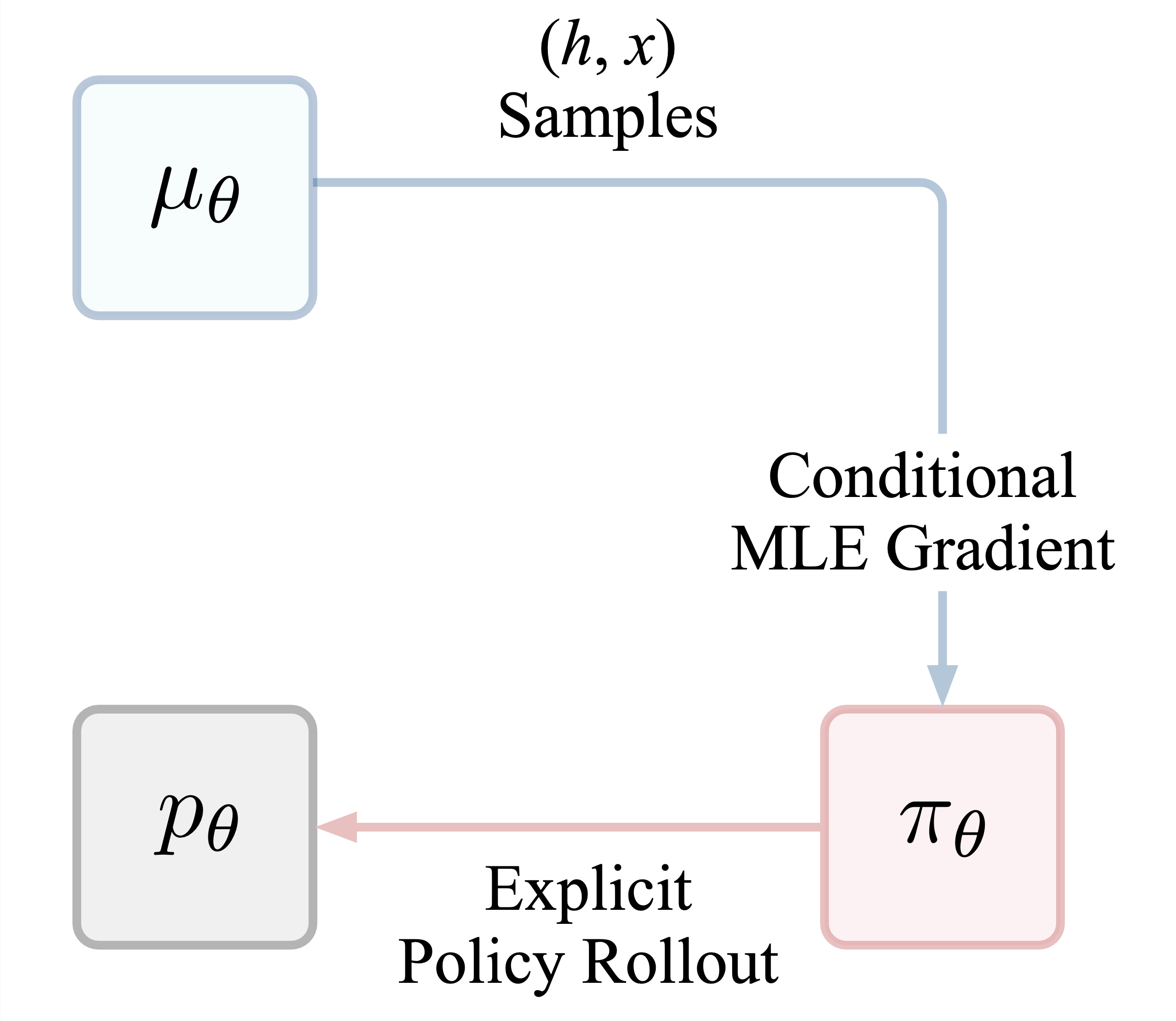}}
\hfill
\subfloat[C-RNN-GAN \cite{mogren2016c}]
{\includegraphics[width=0.26\linewidth]{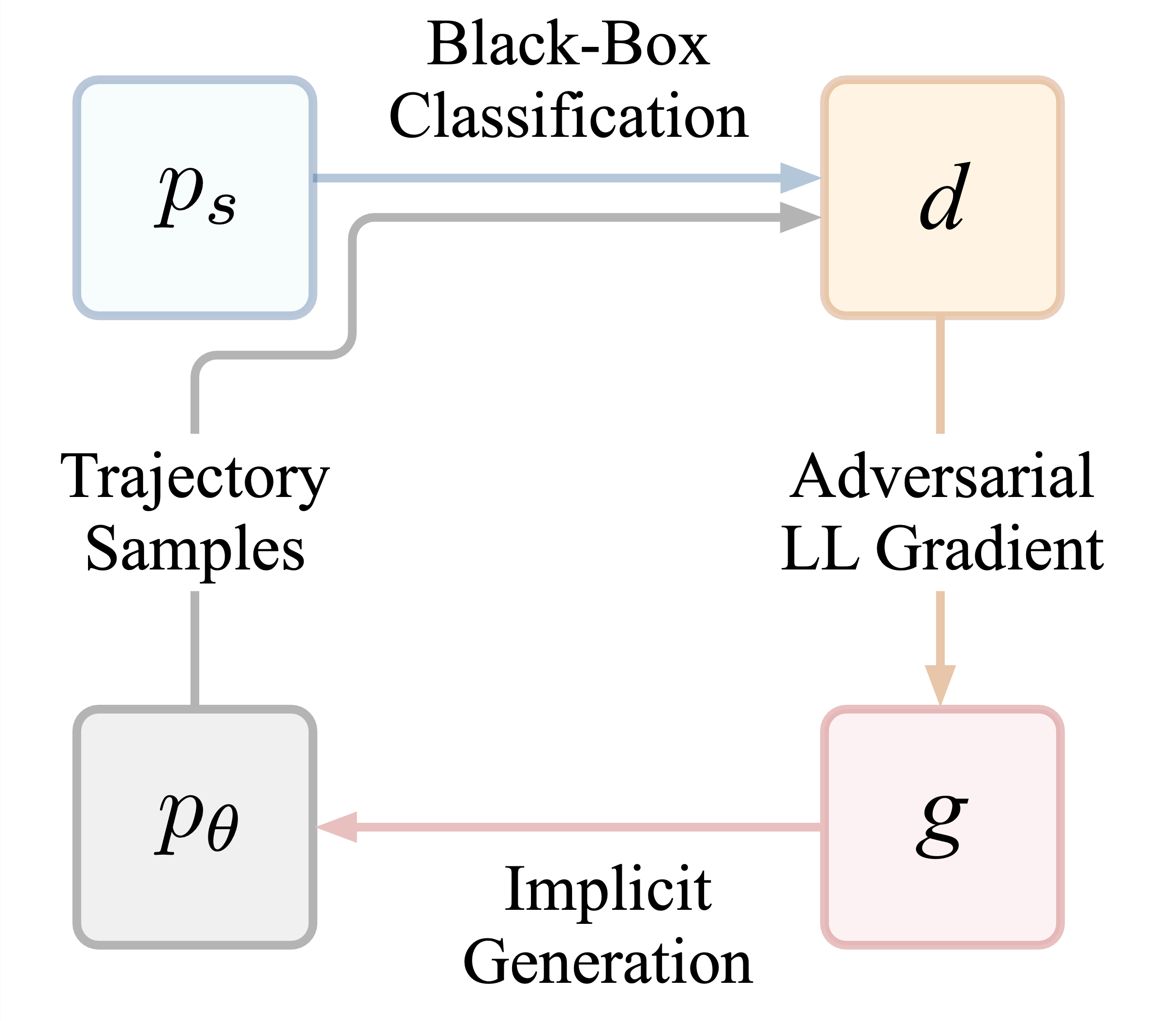}}
\hfill
\subfloat[TimeGAN \cite{yoon2019time}]
{\includegraphics[width=0.26\linewidth]{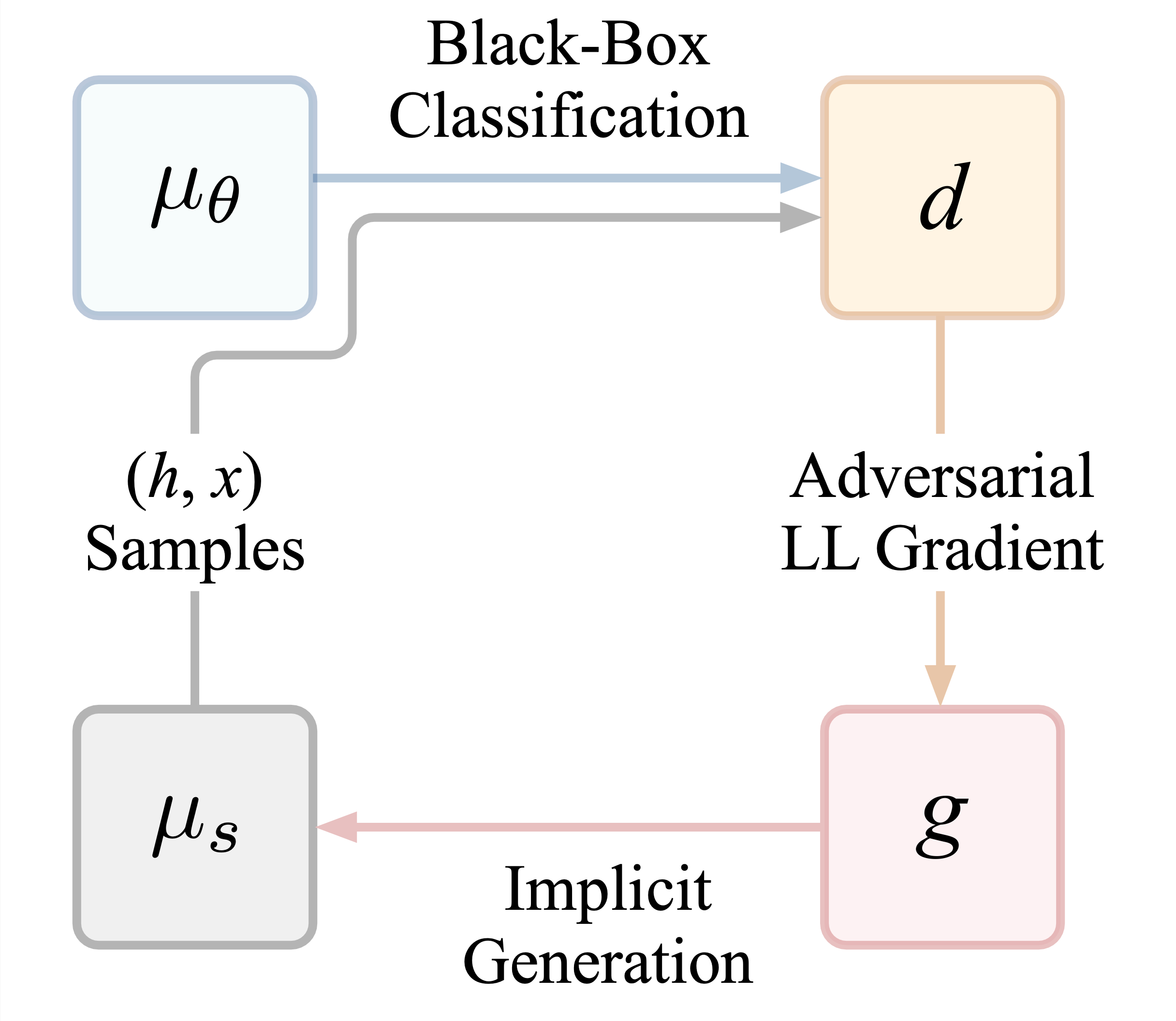}}
\hfill
\subfloat[\textbf{TimeGCI (Ours)}]
{\includegraphics[width=0.26\linewidth]{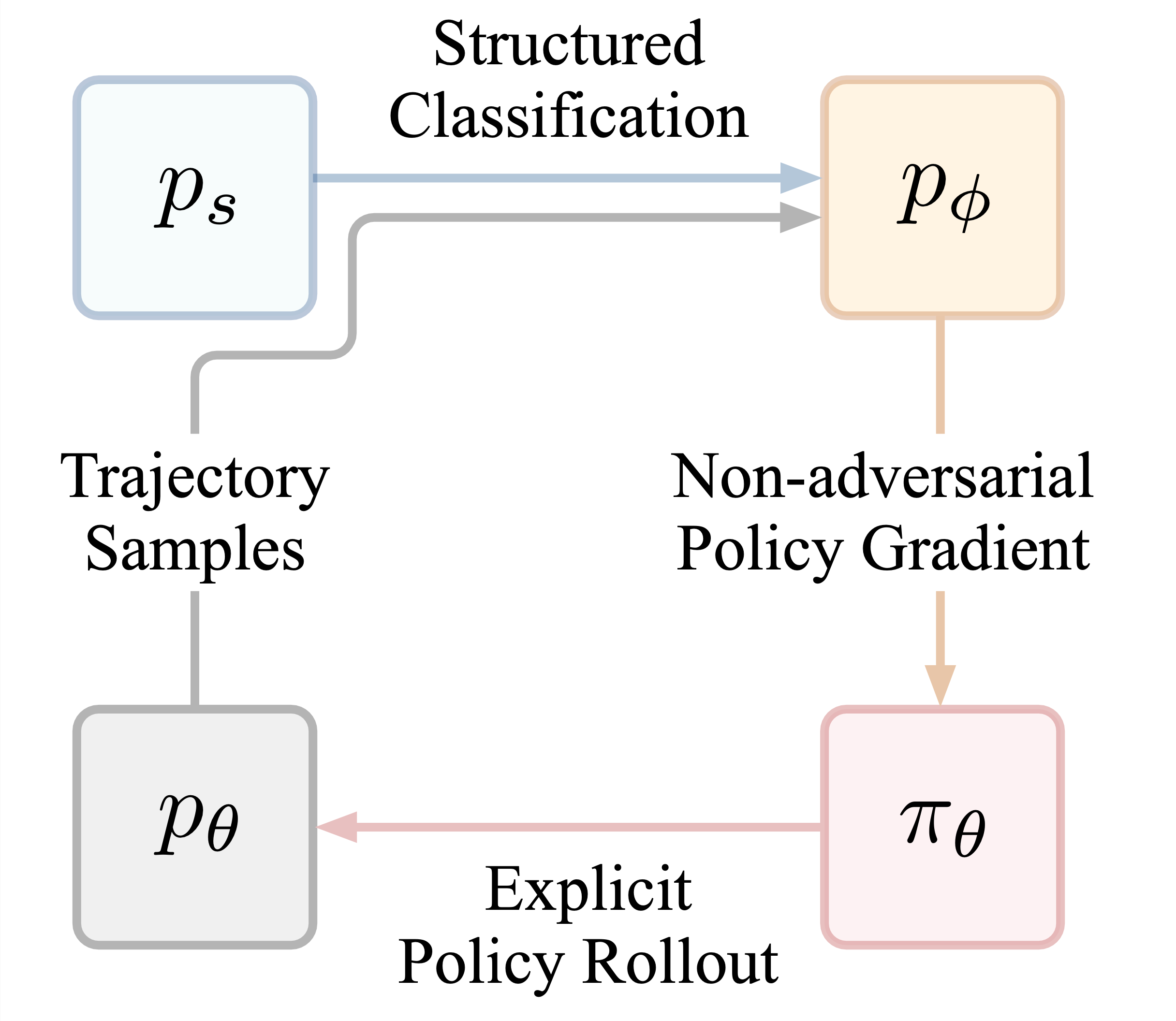}}
}

\caption{\small\dayum{\textit{Comparison of Time-series Generative Models}. Examples of (a) conditional MLE-based autoregressive model, (b) trajectory-centric GAN, and (c) transition-centric GAN. (d) Our proposed technique. See also Table \ref{tab:related}.}}
\label{fig:related}

\vspace{-0.25em}

\end{figure}

\begin{reproposition}[restate=thmopt,name=Global Optimality]\upshape\label{thm:opt}
\dayum{
Let $f_{\phi}\in\mathbb{R}^{\mathcal{H}\times\mathcal{X}}$, and let $p_{\theta}\in\Delta(\mathcal{T})$ be any distribution satisfying positivity: $p_{s}(\tau)>0\Rightarrow p_{\theta}(\tau)>0$ (this does not require $\pi_{\theta}$ be optimal for $f_{\phi}$). Then $\mathcal{L}_{\text{energy}}(\phi;\theta)$ is globally minimized at $F_{\phi}(\cdot)-\log Z_{\phi}=\log p_{s}(\cdot)$, whence $p_{\phi}$ is self-normalized with unit integral.}
\end{reproposition}

\vspace{-0.25em}

\textit{Proof}. Appendix \ref{app:a}. \QED

\vspace{0.25em}

\dayum{
This result is intuitive by analogy with noise-contrastive estimation \cite{gutmann2012noise,goodfellow2015distinguishability}: $\phi$ is learnable as long as negative samples $\tau\sim p_{\theta}$ cover the support of the true $p_{s}$. The positivity condition is mild (e.g. take Gaussian policies $\pi_{\theta}$), and so is the realizability condition (e.g. take neural-networks for $f_{\phi}$). Importantly, note that at optimality classifier $d_{\theta,\phi}$ is \textit{decoupled} from any specific value of $\theta$; contrast this with generic discriminators $d$ in GANs, which are only ever optimal for the current generator.
Now, in practice we must approximate $p_{s}$ and $p_{\theta}$ using \textit{finite} samples. In light of this, two questions are immediate: First, does the learned $\phi$ converge to the global optimum as the sample size increases? Second, what role does the ``quality'' of the policy's samples play in how $\phi$ is learned? For the former:}

\begin{reproposition}[restate=thmcon,name=Asymptotic Consistency]\upshape\label{thm:con}
\dayum{
Let $\phi^{*}$ denote the minimizer for \smash{$\mathcal{L}_{\text{energy}}(\phi;\theta)$}, and let \smash{$\hat{\phi}_{M}^{*}$} denote the minimizer for its finite-data approximation---that is, where the expectations over $p_{s}$ and $p_{\theta}$ are approximated by $M$ samples. Then under some mild conditions, as $M$ increases \smash{$\hat{\phi}_{M}^{*}\overset{_p}{\longrightarrow}\phi^{*}$}.}
\end{reproposition}

\vspace{-0.25em}

\textit{Proof}. Appendix \ref{app:a}. \QED

\vspace{0.25em}

\dayum{
Now for the second question: Clearly if $p_{\theta}$ were too far from $p_{s}$, learning would be slow---the job would be too easy for the classifier $d_{\theta,\phi}$, and it may be able to distinguish samples via basic statistics alone. Indeed, in standard noise-contrastive estimation with a \textit{fixed} noise distribution, learning is ineffective in the presence of many variables \cite{goodfellow2016deep}.
Precisely, however, that is why we continuously update the policy itself as an \textit{adaptive} noise distribution: As $p_{\phi}$ moves closer to $p_{s}$, so does $p_{\theta}$---thus providing more ``challenging'' negative samples.\footnote{It is easy to see that minimizing Equation \ref{eq:ncepolicy} equivalently minimizes the reverse KL div. between $p_{\phi}$ and $p_{\theta}$.} In fact, should we insist on greedily taking each policy update to optimality, we recover a ``weighted'' version of the original max-min gradient from before:}

\begin{reproposition}[restate=thmsce,name=Gradient Equality]\upshape\label{thm:sce}
Let $\phi_{k}$ be the value taken by $\phi$ after the $k$-th gradient update, and let $\theta_{k}^{*}$ denote the associated minimizer for \smash{$\mathcal{L}_{\text{policy}}(\theta;\phi_{k})$}. Suppose $p_{\phi}$ is already normalized; then
\begin{equation*}
\nabla_{\phi}\mathcal{L}_{\text{energy}}(\phi;\theta_{k}^{*})
=
-\tfrac{T}{2}
\nabla_{\phi}\mathcal{L}(\theta_{k}^{*},\phi)
\end{equation*}
\dayum{%
That is, at $\theta_{k}^{*}$ the energy gradient (of Equation \ref{eq:nceenergy}) recovers the original gradient (from Equation \ref{eq:vebmenergy}).
In the general case, suppose $p_{\phi}$ is un-normalized, such that \smash{$\raisebox{1pt}{$p_{\theta_{k}^{*}}$}=p_{\phi}/K_{\phi}$} for some constant $K_{\phi}$; then}

\vspace{-0.50em}

\begin{equation*}
\nabla_{\phi}\mathcal{L}_{\text{energy}}(\phi;\theta_{k}^{*})
=
\tfrac{TK_{\phi}}{K_{\phi}+1}
\mathbb{E}\hspace{-2pt}\raisebox{3pt}{$_{\substack{
h\sim\mu_{\theta_{k}^{*}}\\
x\sim\pi_{\theta_{k}^{*}}(\cdot|h)
}}$}\hspace{-3pt}
\nabla_{\phi}\textstyle f_{\phi}(h,x)
-
\tfrac{T}{K_{\phi}+1}
\mathbb{E}\hspace{-2pt}_{\substack{
h\sim\mu_{s}\\
x\sim\pi_{s}(\cdot|h)
}}\hspace{-4pt}
\nabla_{\phi}\textstyle f_{\phi}(h,x)
\end{equation*}
\end{reproposition}

\vspace{-1.25em}

\textit{Proof}. Appendix \ref{app:a}. \QED

\vspace{0.25em}

\dayum{This ``weighting'' is intuitive: If $p_{\phi}$ were un-normalized such that $K_{\phi}>1$, the energy loss automatically places higher weights on negative samples $h\sim\raisebox{1pt}{$\mu_{\theta_{k}^{*}}$},x\sim\raisebox{1pt}{$\pi_{\theta_{k}^{*}}$}(\cdot|h)$ to bring it down; conversely, if $p_{\phi}$ were un-normalized such that $K_{\phi}<1$, the energy loss places higher weights on positive samples $h\sim\mu_{s},x\sim\pi_{s}(\cdot|h)$ to bring it up. (If $p_{\phi}$ were normalized, then $K_{\phi}=1$ and the weights are equal).
In sum, we have arrived at a framework that learns an explicit sampling policy without exposure bias, a decoupled energy model without nested or saddle-point optimization, and is self-normalizing without importance sampling or estimating the partition function. Figure \ref{fig:related} gives a representative comparison.}

\vspace{-0.15em}

\subsection{Optimization Algorithm}\label{sec:33}

\vspace{-0.05em}

\begin{algorithm}[h!]\small
\captionsetup{font=small}
\algsetup{linenosize=\small}
\begin{spacing}{1.0}
\begin{algorithmic}[1]
\STATE\textbf{Input}: source dataset $\mathcal{D}\approx p_{s}$, mini-batch size $M$, regularization coefficient $\kappa$, learning rates $\lambda$
\STATE\textbf{Initialize}: replay buffer $\mathcal{B}$, energy parameter $\phi$, policy parameter $\theta$, critic parameter $\psi$
\FOR{each iteration}
\FOR{each policy rollout}
\STATE $\mathcal{B}\leftarrow\mathcal{B}\cup\{\tau\sim p_{\theta}\}$
\COMMENT{Generate sample}
\ENDFOR
\FOR{each gradient step}
\STATE $
\asif{\theta}{\psi}
\leftarrow
\asif{\theta}{\psi}
-
\asif{\lambda_{\text{actor}}}{\lambda_{\text{energy}}}
\asif{\nabla_{\theta}}{\nabla_{\psi}}
\asif{\mathcal{L}_{\text{actor}}}{\mathcal{L}_{\text{energy}}}
(
\theta;\phi,\psi
)
+
\kappa
\nabla_{\theta}\mathcal{L}_{\text{mle}}(\theta)
$
\COMMENT{Update policy}
\STATE $
\asif{\phi}{\psi}
\leftarrow
\asif{\phi}{\psi}
-
\asif{\lambda_{\text{energy}}}{\lambda_{\text{energy}}}
\asif{\nabla_{\phi}}{\nabla_{\psi}}
\asif{\mathcal{L}_{\text{energy}}}{\mathcal{L}_{\text{energy}}}
(
\phi;\theta
)
$
\COMMENT{Update energy}
\STATE $
\psi
\leftarrow
\psi
-
\asif{\lambda_{\text{critic}}}{\lambda_{\text{energy}}}
\nabla_{\psi}
\asif{\mathcal{L}_{\text{critic}}}{\mathcal{L}_{\text{energy}}}
(
\psi;\phi
)
$
\COMMENT{Update critic}
\ENDFOR
\ENDFOR
\STATE\textbf{Output}: learned policy parameter $\theta^{*}$ and energy parameter $\phi^{*}$
\end{algorithmic}
\end{spacing}
\caption{~Time-series Generation by Contrastive Imitation\COM{Details in Appendix \ref{app:b}}}\label{alg:timegci}
\end{algorithm}

\dayum{
The only remaining choice is the method of policy optimization. Here we employ \textit{soft actor-critic} \cite{haarnoja2018soft}, although in principle any technique  will do---the only requirement is that it performs reinforcement learning with \textit{entropy-regularization} \cite{fox2016taming,haarnoja2017reinforcement,shi2019soft}. To optimize the policy per Equation \ref{eq:ncepolicy}, in addition to the policy ``actor'' itself, this trains a ``critic'' to estimate value functions. As usual, the actor takes soft policy improvement steps, minimizing \smash{$
\mathcal{L}_{\text{actor}}(\theta;\phi,\psi)
\coloneqq
\mathbb{E}_{h\sim\mathcal{B}}~
\mathbb{E}_{x\sim\pi_{\theta}(\cdot|h)}
[\log\pi_{\theta}(x|h)-Q_{\psi}(h,x)]
$}, where $Q_{\psi}:\mathcal{H}\times\mathcal{X}\rightarrow\mathbb{R}$ is the transition-wise soft value function parameterized by $\psi$, and $\mathcal{B}$ is a replay buffer of samples generated by $\pi_{\theta}$.
For stability, the actor is regularized with the conditional MLE loss \smash{$\mathcal{L}_{\text{mle}}(\theta)\coloneqq\mathbb{E}_{x\sim\pi_{s}(\cdot|h)}\log\pi_{\theta}(x|h)$}.
The critic is trained to minimize the soft Bellman residual:
\smash{$
\mathcal{L}_{\text{critic}}(\psi;\phi)
\coloneqq
\mathbb{E}_{h,x\sim\mathcal{B}}
(
Q_{\psi}(h,x)-f_{\phi}(h,x)-V_{\psi}(h')
)^{2}
$}, where the state-values are bootstrapped as $V_{\psi}(h')\coloneqq\mathbb{E}_{x'\sim\pi_{\theta}(\cdot|h')}
[Q_{\psi}(h',x')-\log\pi_{\theta}(x'|h')]$. By expressly training an \textit{imitation} policy to mimic time-series behavior using rewards from an energy model trained by \textit{contrastive} learning, we call this framework Time-series Generation by Contrastive Imitation (TimeGCI): See Algorithm \ref{alg:timegci}.}

\vspace{-0.10em}

\section{Discussion}\label{sec:4}

\vspace{-0.10em}

Our theoretical motivations are apparent (Sections \ref{sec:22}--\ref{sec:31}), and the practical mechanics of optimization are straightforward (Section \ref{sec:32}--\ref{sec:33}). To understand the strengths and limitations of TimeGCI, two questions remain: First, how does this relate to bread-and-butter imitation learning of sequential decision-making? Second, how does this compare with recent deep generative models for time series?

\begin{table*}[t]\small
\newcolumntype{A}{>{\centering\arraybackslash}m{0.7 cm}}
\newcolumntype{B}{>{\centering\arraybackslash}m{2.7 cm}}
\newcolumntype{C}{>{\centering\arraybackslash}m{4.3 cm}}
\newcolumntype{D}{>{\centering\arraybackslash}m{1.3 cm}}
\newcolumntype{E}{>{\centering\arraybackslash}m{1.3 cm}}
\newcolumntype{F}{>{\centering\arraybackslash}m{1.5 cm}}
\newcolumntype{G}{>{\centering\arraybackslash}m{1.5 cm}}
\newcolumntype{H}{>{\centering\arraybackslash}m{1.4 cm}}
\newcolumntype{I}{>{\centering\arraybackslash}m{1.3 cm}}
\newcolumntype{J}{>{\centering\arraybackslash}m{1.1 cm}}
\setlength{\cmidrulewidth}{0.5pt}
\setlength\tabcolsep{0pt}
\renewcommand{\arraystretch}{1.06}
\caption{\small\dayum{\textit{Comparison of Time-series Generative Models}. Examples of conditional MLE-based autoregressive models, trajectory-centric GANs, transition-centric GANs, as well as our proposed technique. See also Figure \ref{fig:related}.}}

\vspace{-1.00em}

\label{tab:related}
\begin{center}
\begin{adjustbox}{max width=\textwidth}
\begin{tabular}{ABCDEFGHIJ}
\toprule
  \textbf{Type}
& \textbf{Examples}
& \textbf{Optimization~Objective(s)}
& Generator Signal
& Discrim. Signal
& No Ex- posure Bias
& Decoupled Discrim.
& Non-Adversarial
& Explicit Policy
& Explicit Energy
\\
\midrule
  \multirow{3}{*}{\rotatebox[origin=c]{90}{\makecell{Condit.\\MLE}}}
& T-Forcing \cite{williams1989learning}
& Data LL
& Stepwise
& \na
& \xm
& \na
& \cm
& \cm
& \xm
\\
& Z-Forcing \cite{goyal2017z}
& Data LL (ELBO)
& Stepwise
& \na
& \xm
& \na
& \cm
& \cm
& \xm
\\
& P-Forcing \cite{lamb2016professor}
& Data LL + Class. LL ($p_{\theta}$ v. $\tilde{p}_{\theta}$)
& Stepwise
& Global
& \xm
& \xm
& \xm
& \cm
& \xm
\\
\midrule
  \multirow{3}{*}{\rotatebox[origin=c]{90}{\makecell{Traject.\\GAN}}}
& C-RNN-GAN \cite{mogren2016c}
& Classification LL ($p_{\theta}$ v. $p_{s}$)
& Global
& Global
& \cm
& \xm
& \xm
& \xm
& \xm
\\
& DoppelGANger \cite{lin2019generating}
& Classification LL ($p_{\theta}$ v. $p_{s}$)
& Global
& Global
& \cm
& \xm
& \xm
& \xm
& \xm
\\
& COT-GAN \cite{xu2020cot}
& Sinkhorn Divergence ($p_{\theta}$ v. $p_{s}$)
& Global
& Global
& \cm
& \xm
& \xm
& \xm
& \xm
\\
\midrule
  \multirow{3}{*}{\rotatebox[origin=c]{90}{\makecell{Transit.\\GAN}}}
& RC-GAN \cite{esteban2017real}
& Classification LL ($\mu_{\theta}$ v. $\mu_{s}$)
& Stepwise
& Stepwise
& \cm
& \xm
& \xm
& \xm
& \xm
\\
& T-CGAN \cite{ramponi2018t}
& Classification LL ($\mu_{\theta}$ v. $\mu_{s}$)
& Stepwise
& Stepwise
& \cm
& \xm
& \xm
& \xm
& \xm
\\
& TimeGAN \cite{yoon2019time}
& Class. LL ($\mu_{\theta}$ v. $\mu_{s}$) + Data LL
& Stepwise
& Stepwise
& \cm
& \xm
& \xm
& \xm
& \xm
\\
\midrule
  \multicolumn{2}{c}{\textbf{TimeGCI (Ours)}}
& Discrim.: Class. LL ($p_{\theta}$ v. $p_{s}$) Generator: Policy Optimization
& Stepwise
& Global
& \cm
& \cm
& \cm
& \cm
& \cm
\\
\bottomrule
\end{tabular}
\end{adjustbox}
\end{center}

\vspace{-1.25em}

\end{table*}

\dayum{
\textbf{Imitation Perspective}~
In sequential decision-making, \textit{imitation learning} deals with training a policy purely on the basis of demonstrated behavior---that is, with no knowledge of the reward signals that induced the behavior in the first place \cite{ross2011reduction,osa2018imitation,attia2018global}. Consider the standard Markov decision process setting, with states $z\in\mathcal{Z}$, actions $u\in\mathcal{U}$, dynamics $\omega\in\Delta(\mathcal{Z})^{\mathcal{Z}\times\mathcal{U}}$, and rewards $\rho\in\mathbb{R}^{\mathcal{Z}\times\mathcal{U}}$. Classically, imitation learning seeks to minimize the regret \smash{$\mathcal{R}_{s}(\theta)$\pix$\coloneqq$\pix$\mathbb{E}_{\pi_{s}}[\smallsum_{t}\rho(z_{t},u_{t})]$$-$$\mathbb{E}_{\pi_{\theta}}[\smallsum_{t}\rho(z_{t},u_{t})]$}, with $\pi_{s},\pi_{\theta}\in\Delta(\mathcal{U})^{\mathcal{Z}}$ here being the demonstrator and imitator policies, and expectations are taken over episodes generated per $u_{t}\sim\pi(\cdot|z_{t})$ and $z_{t+1}\sim\omega(\cdot|z_{t},u_{t})$ \cite{ross2010efficient,syed2010reduction}. First, observe that by interpreting $h$ as ``states'' and $x$ as ``actions'', our problem setup bears a precise resemblance to imitation learning:}

\vspace{-0.25em}

\begin{recorollary}[restate=thmreduce,name=Generation as Imitation]\upshape\label{thm:reduce}
\dayum{
Let state space $\mathcal{Z}\coloneqq\mathcal{H}$, action space $\mathcal{U}\coloneqq\mathcal{X}$, and reward function $\rho\coloneqq f_{s}$. In addition, let the dynamics be such that $\omega(\cdot|h_{t},x_{t})$ is the Dirac delta centered at \smash{$h_{t+1}$$\coloneqq$\pix$(x_{1},...,x_{t})$}. Then the regret exactly corresponds to the expected quality difference: $\mathcal{R}_{s}$\pix$=$\pix$\Delta\bar{F}_{s}$.}
\end{recorollary}

\vspace{-0.25em}

\textit{Proof}. Immediate from Definition \ref{def:difference}. \QED

\vspace{0.25em}

\dayum{
Now, since we want low regret but have no knowledge of the true quality measure (i.e. ``reward sign-
al''), we may naturally learn it together. In this sense, TimeGCI is analogous to imitation by \textit{inverse reinforcement learning} (IRL), which seeks to infer rewards that plausibly induced the demonstrated behavior, and to optimize imitating policies on that basis \cite{ng2000algorithms,abbeel2004apprenticeship,bica2021learning,jarrett2021inverse}.
Further, in simultaneously optimizing for variety (cf. entropy) and typicality (cf. energy), TimeGCI is analogous to maximum-entropy IRL \cite{ziebart2008maximum,finn2016guided}. Our contrastive approach also bears mild resemblance to stepwise discriminators studied in this vein \cite{fu2018learning,qureshi2019adversarial}, although our framework focuses on trajectory-wise modeling, and is not adversarial (see Appendix \ref{app:d} for more discussion on how TimeGCI relates to popular imitation learning methods).}

\dayum{
There are also crucial differences: In imitation learning, dynamics are generally Markovian; states are readily defined as discrete elements or real vectors, and action spaces are small/discrete. The practical challenge is sample efficiency---to reduce the cost of environment interactions \cite{kostrikov2019discriminator,blonde2019sample}. In time-series generation, however, rollouts are free---generating a synthetic trajectory does not require interacting with the real world. But dynamics are never Markovian: The practical challenge is that representations of variable-length histories must be jointly learned. Moreover, actions are the full-dimensional feature vectors themselves, which renders policy optimization more demanding than usual (see Appendix \ref{app:b}); beyond the tractable tabular settings we experiment in, higher-dimensional data may prove challenging.} 

%
%
%
%

\dayum{
\textbf{Related Work}~
Table \ref{tab:related} summarizes the key differentiators of TimeGCI from prevailing techniques. As discussed in Section \ref{sec:1}, MLE-based autoregressive models \cite{williams1989learning,goyal2017z,lamb2016professor} are easy to optimize, and learn explicit conditional distributions that can be used for inspection, resampling, or uncertainty estimation,
but they suffer from exposure bias \cite{bengio2009curriculum,huszar2016not,yoon2019time}. GAN-based adversarial models fall into two camps: For trajectory-centric methods \cite{mogren2016c,lin2019generating,xu2012distributionally}, with only sequence-level signals to guide the generator, they often struggles to converge to the adversarial objective without extensive tuning \cite{yoon2019time}---with the exception of \cite{xu2012distributionally}, which utilizes Sinkhorn divergences instead. Transition-centric methods \cite{esteban2017real,ramponi2018t,yoon2019time} provide more granular signals to guide the generator, but this simply alters the objective of learning $p_{s}$ to one of learning $\mu_{s}$, and still inherits the disadvantages of implicit, adversarial learning.}

\dayum{
Our analysis is built on ideas from energy-based models (EBMs) \cite{lecun2006tutorial,xie2016theory,du2019implicit} and reinforcement learning for sequence prediction \cite{bachman2015data,venkatraman2015improving,keneshloo2019deep}. In particular, our initial formulation (Section \ref{sec:31}) can be viewed as a temporal extension of variational EBMs \cite{kim2016deep,zhai2016generative}. Moreover, by adaptively learning $\pi_{\theta}$ to give negative samples for $d_{\theta,\phi}$, the formulation we study (Section \ref{sec:32}) is equivalent to a temporal analogue of noise-contrastive estimation (NCE) \cite{gutmann2010noise,goodfellow2015distinguishability}. More tangentially, conditional EBMs have been trained with NCE for text generation \cite{mnih2012fast,mnih2013learning,ma2018noise}, and the strength of global normalization has been studied \cite{andor2016globally}; that said, these are confined to the case where external input tokens are available for conditioning at each step---and not free-running as in our time-series setting. Finally, note that viewing sequence generation as a decision-making problem is present in language modeling \cite{ranzato2016sequence,bahdanau2017actor} where task-specific metrics are available as signals. In the absence of predefined signals, GAN-based methods that jointly train discriminators to provide rewards for imitation have been studied \cite{yu2017seqgan,lu2019cot,yin2020meta,fedus2018maskgan,guo2018long,zhang2017adversarial}, although they are adversarial, and all focus on the special case of generating discrete tokens for language modeling.}

\vspace{-0.75em}

\section{Experiments}\label{sec:5}

\vspace{-0.35em}

\dayum{
\textbf{Benchmarks}~
We test Algorithm \ref{alg:timegci} (\textbf{TimeGCI}) against the following:
The classic Teacher Forcing trains autoregressive networks using ground-truth conditioning (\textbf{T-Forcing}) \cite{williams1989learning}.
Professor Forcing uses adversarial domain adaptation by training an auxiliary discriminator to encourage dynamics of the network's free-running and teacher-forced states to be similar (\textbf{P-Forcing}) \cite{lamb2016professor}.
Trajectory-centric recurrent GANs (\textbf{C-RNN-GAN}) directly plug RNNs into the GAN framework as generators and discriminators for full sequences \cite{mogren2016c}.
Causal Optimal Transport GAN (\textbf{COT-GAN}) is the latest variant of this  \cite{xu2020cot}, proposing to approximate Sinkkorn divergences instead of the standard JS divergence.
For transition-centric recurrent GANs (\textbf{RC-GAN}), the adversarial loss is computed as the sum of log likelihoods for the stepwise feature vectors conditioned on histories \cite{esteban2017real}, instead of directly as the log likelihood for the entire sequence.
Finally, Time-series GAN (\textbf{TimeGAN}) is its latest incarnation \cite{yoon2019time}, proposing to generate and discriminate within a jointly optimized embedding space for efficiency.}

\begin{wraptable}{r}{0.42\textwidth}\small
\renewcommand{\arraystretch}{1.00}
\setlength\tabcolsep{2pt}
\centering

\vspace{-1.35em}

\caption{\textit{Summary Statistics for Datasets Used}.}
\label{tab:datasets}

\vspace{-1.00em}

\begin{center}
\resizebox{0.4225\textwidth}{!}{%
\begin{tabular}{l|ccccc}
\toprule
\textit{Dataset} & Dimension & Length & Autocor. & +3 Lag & +5 Lag \\
\midrule
Sines     &  5 & 24 & 0.875 & 0.623 & 0.377 \\
Metro     &  9 & 24 & 0.429 & 0.200 & 0.029 \\
Gas       & 20 & 24 & 0.656 & 0.382 & 0.170 \\
Energy    & 29 & 24 & 0.702 & 0.411 & 0.176 \\
MIMIC-III & 52 & 24 & 0.532 & 0.212 & 0.059 \\
\bottomrule
\end{tabular}
}
\end{center}

\vspace{-1.25em}

\end{wraptable}

\dayum{
\textbf{Datasets}~
We employ five tabular time-series datasets with a variety of different characteristics, such as periodicity, noise level, and correlations:
First, we use a synthetic dataset of multivariate sinusoids with different frequencies and phases (\textbf{Sines}) \cite{yoon2019time}.
Second, we use a UCI dataset from the monitored energy usage of household appliances in a low-energy house (\textbf{Energy}) \cite{candanedo2017data}.
Third, we use a UCI dataset from temperature-modulated semiconductor gas sensors for chemical detection (\textbf{Gas}) \cite{burgues2018estimation}.
Fourth, we use a UCI dataset of hourly interstate vehicle volume at a state traffic recording station (\textbf{Metro}) \cite{hogue2018data}.
Fifth, we use a medical dataset of intensive-care patients from the Medical Information Mart for Intensive Care (\textbf{MIMIC-III}) \cite{johnson2016mimic}. All datasets are accessible from their sources, and we use the original source code for preprocessing sines and the UCI datasets by \cite{yoon2019time}, publicly available at \cite{yoon2019github}. Table \ref{tab:datasets} shows summary statistics for the datasets used in the experiments.}

\dayum{
\textbf{Implementation}~
Experiments for each dataset are arranged as follows: The real trajectories that constitute the original dataset $\mathcal{D}$ are fed as input to train all algorithms. Each algorithm is subsequently used in test mode to generate 10,000 synthetic trajectories. Then, the performance of each algorithm is evaluated on the basis of these generated trajectories. This process is then performed for a total of 10 repetitions, from which we compile the means and standard errors for each reported result. For fair comparison, analogous network components across all benchmarks share the same recurrent architecture: Wherever a generator, policy, discriminator, energy, or critic network applies, we use LSTMs with one hidden layer of 32 units to compute hidden states for representing histories $h$, and two fully-connected hidden layers of 32 units each and ELU activations to compute task-specific output variables (i.e. the generator output, policy parameters, discriminator output, energy functions, or critic values). In other respects, we use the publicly available source code to construct the benchmark algorithms---accessible at \cite{yoon2019github,xu2020github,mogren2016github,esteban2017github,lamb2016github}. See Appendix \ref{app:c} for additional detail on hyperparameters and implementations.}

\begin{table*}[h!]\small
\renewcommand{\arraystretch}{0.97}
\setlength\tabcolsep{8pt}
\centering
\caption{\textit{Performance Comparison of TimeGCI and Benchmarks}. Bold numbers indicate best-performing results.}

\vspace{-1.00em}

\label{tab:result}
\begin{center}
\resizebox{\textwidth}{!}{%
\begin{tabular}{l|l|ccccc}
\toprule
\textit{Benchmark} & \textit{Metric} & Sines & Energy & Gas & Metro & MIMIC-III \\
\midrule
  \multirow{4}{*}{T-Forcing}
& Predictive Score
& 0.108 $\pm$ 0.002 & 0.310 $\pm$ 0.001 & 0.035 $\pm$ 0.003 & 0.242 $\pm$ 0.001 & 0.017 $\pm$ 0.001 \\
& +3 Steps Ahead
& 0.115 $\pm$ 0.001 & 0.281 $\pm$ 0.001 & 0.080 $\pm$ 0.001 & 0.244 $\pm$ 0.001 & 0.024 $\pm$ 0.007 \\
& +5 Steps Ahead
& 0.122 $\pm$ 0.003 & 0.270 $\pm$ 0.002 & 0.111 $\pm$ 0.001 & 0.248 $\pm$ 0.001 & 0.018 $\pm$ 0.003 \\
& $x$-Corr. Score
& 8.369 $\pm$ 0.015 & 194.1 $\pm$ 0.043 & 150.8 $\pm$ 0.067 & 4.222 $\pm$ 0.013 & 400.9 $\pm$ 3.203 \\
\midrule
  \multirow{4}{*}{P-Forcing}
& Predictive Score
& 0.105 $\pm$ 0.001 & 0.303 $\pm$ 0.002 & 0.037 $\pm$ 0.001 & 0.241 $\pm$ 0.001 & 0.023 $\pm$ 0.006 \\
& +3 Steps Ahead
& 0.110 $\pm$ 0.001 & 0.268 $\pm$ 0.002 & 0.086 $\pm$ 0.002 & 0.241 $\pm$ 0.001 & 0.018 $\pm$ 0.001 \\
& +5 Steps Ahead
& 0.115 $\pm$ 0.001 & 0.259 $\pm$ 0.002 & 0.121 $\pm$ 0.002 & 0.242 $\pm$ 0.001 & 0.017 $\pm$ 0.001 \\
& $x$-Corr. Score
& 8.156 $\pm$ 0.010 & 207.6 $\pm$ 0.057 & 150.5 $\pm$ 0.023 & 3.014 $\pm$ 0.006 & 346.6 $\pm$ 2.901 \\
\midrule
  \multirow{4}{*}{C-RNN-GAN}
& Predictive Score
& 0.751 $\pm$ 0.001 & 0.500 $\pm$ 0.001 & 0.242 $\pm$ 0.001 & 0.419 $\pm$ 0.005 & 0.019 $\pm$ 0.001 \\
& +3 Steps Ahead
& 0.769 $\pm$ 0.001 & 0.500 $\pm$ 0.001 & 0.243 $\pm$ 0.001 & 0.416 $\pm$ 0.002 & 0.020 $\pm$ 0.001 \\
& +5 Steps Ahead
& 0.786 $\pm$ 0.001 & 0.501 $\pm$ 0.001 & 0.241 $\pm$ 0.001 & 0.416 $\pm$ 0.003 & 0.019 $\pm$ 0.001 \\
& $x$-Corr. Score
& 10.76 $\pm$ 0.012 & 644.2 $\pm$ 0.112 & 266.4 $\pm$ 0.008 & 18.39 $\pm$ 0.003 & 1720. $\pm$ 0.339 \\
\midrule
  \multirow{4}{*}{COT-GAN}
& Predictive Score
& 0.099 $\pm$ 0.001 & 0.259 $\pm$ 0.001 & 0.022 $\pm$ 0.001 & 0.245 $\pm$ 0.001 & 0.014 $\pm$ 0.001 \\
& +3 Steps Ahead
& 0.109 $\pm$ 0.001 & 0.261 $\pm$ 0.001 & 0.050 $\pm$ 0.001 & 0.246 $\pm$ 0.001 & 0.013 $\pm$ 0.001 \\
& +5 Steps Ahead
& 0.110 $\pm$ 0.001 & 0.262 $\pm$ 0.001 & 0.072 $\pm$ 0.001 & 0.245 $\pm$ 0.001 & 0.013 $\pm$ 0.001 \\
& $x$-Corr. Score
& 3.114 $\pm$ 0.038 & \textbf{67.93} $\pm$ \textbf{0.227} & \textbf{25.56} $\pm$ \textbf{0.156} & 3.055 $\pm$ 0.013 & 497.7 $\pm$ 2.581 \\
\midrule
  \multirow{4}{*}{RC-GAN}
& Predictive Score
& 0.751 $\pm$ 0.001 & 0.498 $\pm$ 0.001 & 0.243 $\pm$ 0.001 & 0.412 $\pm$ 0.003 & 0.019 $\pm$ 0.001 \\
& +3 Steps Ahead
& 0.770 $\pm$ 0.001 & 0.500 $\pm$ 0.001 & 0.244 $\pm$ 0.001 & 0.415 $\pm$ 0.004 & 0.019 $\pm$ 0.001 \\
& +5 Steps Ahead
& 0.786 $\pm$ 0.001 & 0.499 $\pm$ 0.001 & 0.243 $\pm$ 0.001 & 0.418 $\pm$ 0.004 & 0.018 $\pm$ 0.001 \\
& $x$-Corr. Score
& 5.649 $\pm$ 0.012 & 582.3 $\pm$ 0.047 & 231.2 $\pm$ 0.003 & 19.77 $\pm$ 0.001 & 1592. $\pm$ 0.192 \\
\midrule
  \multirow{4}{*}{TimeGAN}
& Predictive Score
& 0.196 $\pm$ 0.006 & 0.261 $\pm$ 0.001 & 0.264 $\pm$ 0.011 & 0.245 $\pm$ 0.002 & 0.502 $\pm$ 0.023 \\
& +3 Steps Ahead
& 0.223 $\pm$ 0.006 & 0.263 $\pm$ 0.001 & 0.251 $\pm$ 0.014 & 0.243 $\pm$ 0.001 & 0.484 $\pm$ 0.021 \\
& +5 Steps Ahead
& 0.246 $\pm$ 0.005 & 0.262 $\pm$ 0.005 & 0.252 $\pm$ 0.012 & 0.242 $\pm$ 0.001 & 0.453 $\pm$ 0.020 \\
& $x$-Corr. Score
& 17.86 $\pm$ 0.001 & 667.5 $\pm$ 0.001 & 282.5 $\pm$ 0.001 & 17.11 $\pm$ 0.001 & 2140. $\pm$ 0.010 \\
\midrule
  \multirow{4}{*}{\textbf{TimeGCI (Ours)}}
& Predictive Score
& \textbf{0.097} $\pm$ \textbf{0.001} & \textbf{0.251} $\pm$ \textbf{0.001} & \textbf{0.018} $\pm$ \textbf{0.000} & \textbf{0.239} $\pm$ \textbf{0.001} & \textbf{0.002} $\pm$ \textbf{0.000} \\
& +3 Steps Ahead
& \textbf{0.104} $\pm$ \textbf{0.001} & \textbf{0.251} $\pm$ \textbf{0.001} & \textbf{0.042} $\pm$ \textbf{0.001} & \textbf{0.239} $\pm$ \textbf{0.001} & \textbf{0.001} $\pm$ \textbf{0.000} \\
& +5 Steps Ahead
& \textbf{0.109} $\pm$ \textbf{0.001} & \textbf{0.251} $\pm$ \textbf{0.001} & \textbf{0.067} $\pm$ \textbf{0.001} & \textbf{0.239} $\pm$ \textbf{0.001} & \textbf{0.001} $\pm$ \textbf{0.000} \\
& $x$-Corr. Score
& \textbf{1.195} $\pm$ \textbf{0.011} & 105.2 $\pm$ 0.433 & 47.91 $\pm$ 0.811 & \textbf{0.738} $\pm$ \textbf{0.019} & \textbf{194.3} $\pm$ \textbf{0.180} \\
\bottomrule
\end{tabular}
}
\end{center}

\vspace{-2.00em}

\end{table*}

\dayum{
\textbf{Evaluation and Results}~
In the tabular data setting, assessing synthetic data generation is inherently tricky \cite{xu2019modeling,park2018data,alaa2021faithful}: Unlike in media-specific applications, we have no predefined measures such as music polyphony or \textsc{bleu} scores, nor can we use human evaluation of realism as done for videos. For tabular time-series, the generally accepted standard for comparing synthetic data is to apply the \textit{Train-on-Synthetic, Test-on-Real} (TSTR) framework, first proposed by \cite{esteban2017real} and employed by most recent work in synthetic time-series generation \cite{esteban2017real,ramponi2018t,yoon2019time,alaa2020generative,fekri2020generating,tgan_smartgrid}, as well as more generally for tabular synthetic data of any kind \cite{xu2019modeling,park2018data,jordon2019pate}.
Specifically, we apply the performance measure used by \cite{yoon2019time,alaa2020generative,fekri2020generating} to quantify how much the synthetic sequences inherit the predictive characteristics of the original dataset (\textbf{Predictive Score}): Using synthetic samples, a generic post-hoc sequence-prediction model is learned to forecast next-step feature vectors over training sequences. Then, the trained model is evaluated on the original data, and its predictive performance is quantified in terms of the mean absolute error. We use the original source code for computing this metric, publicly available at \cite{yoon2019github}.}

\dayum{
Further to prior works using this measure, we additionally believe that synthetic data evaluation should be more general than just next-step TSTR forecasting. After all, the distinguishing characteristic of sequential (vs. static) data generation is that we care about evolution of features \textit{over time}. Hence we also compute TSTR metrics for horizons of other lengths (\textbf{+3 Steps Ahead} and \textbf{+5 Steps Ahead}). Importantly, note that a key strength of TSTR evaluation is in its sensitivity to \textit{mode collapse}: If any generation scheme suffers from mode collapse (as GAN methods are prone to), TSTR scores would degrade due to the synthetic data failing to capture the diversity of the real data, which means any prediction model trained on that basis would also fail to capture this variation). Finally, similar to some recent works \cite{lin2019generating,xu2020cot}, we also compute the cross-correlations of real and synthetic feature vectors, and report the sum of the absolute differences between them, averaged over time ($\bm{x}$-\textbf{Corr. Score}); this serves to verify if feature relationships are preserved well, in addition to temporal relationships.
Table \ref{tab:result} shows the results: With respect to these metrics, we find that TimeGCI somewhat consistently produces synthetic samples that perform similarly or better than benchmark algorithms in all datasets. (Note that we do empirically observe several instances of mode collapse in GAN-based benchmarks).}

\vspace{-0.10em}

\section{Conclusion}\label{sec:6}

\vspace{-0.10em}

\dayum{
In this work, we invite an explicit analogy between time-series generation and imitation learning, and explore a framework that fleshes out this connection. Two caveats are in order: First, while we began from the notion of moment-matching to address the error compounding problem, in practice there is no guarantee that this is accomplished well during optimization. In particular, \textit{scalability} is a major limitation beyond the range of feature dimensions and sequence lengths considered in our experiments. Sample-based estimates could rapidly degrade with the horizon, especially if transitions are highly stochastic. A relevant question is whether or not training on fixed subsequence lengths could potentially alleviate this concern for longer sequences. In addition, while our approach seeks to dispense with the instabilities typical of adversarial training, we are instead left with the difficulties of policy optimization, which may prove a prohibitive challenge in higher-dimensional feature spaces. For the datasets we consider, we find that pre-training and regularizing the policy with maximum likelihood, combined with a small enough learning rate, had the most impact in promoting stability and learning.
Second, we reiterate that a perennial challenge in modeling tabular data is in choosing the metric for \textit{evaluation}. While we opted for the most commonly accepted method of TSTR, this may not be general enough to capture the range of downstream tasks that may be performed on the synthetic data. Future work will benefit from a deeper investigation into more sophisticated measures for time series, such as contrastive methods and how to evaluate different aspects of the ``quality'' of the generated trajectories.}

\vspace{-0.10em}

\section*{Acknowledgments }

\vspace{-0.10em}

\dayum{
We would like to thank the reviewers for all their invaluable feedback. This work was supported by Alzheimer’s Research UK, The Alan Turing Institute under the EPSRC grant EP/N510129/1, the US Office of Naval Research, as well as the National Science Foundation under grant number 1722516.}

\clearpage
\bibliographystyle{unsrt}
\bibliography{
bib/0-imitate,
bib/1-reinforce,
bib/2-entropy,
bib/3-information,
bib/4-constraints,
bib/5-risk,
bib/6-intrinsic,
bib/7-bounded,
bib/8-identification,
bib/9-interpret,
bib/a-miscellaneous,
bib/b-time,
bib/c-yoon
}

\clearpage
\appendix


\section{Proofs of Propositions}\label{app:a}

\dayum{
For Lemmas \ref{thm:local} and \ref{thm:global}, we first introduce some additional quantities to enable more compact notation; in particular, we adopt the value-function terminology from imitation learning. The following definitions are standard and immediate from the mapping given by Corollary \ref{thm:reduce}, but are explicitly stated here for completeness. Recall that $f_{s}:\mathcal{H}\times\mathcal{X}\rightarrow[-c,c]$ for some finite $c$. At any state $h_{t}$, define the ``value function'' to be the (forward-looking) expected sum of future quantities $f_{s}(h_{u},x_{u})$ for $u=t,...,T$. Specifically, let $V_{s,t}^{\pi_{\theta}}(h):\mathcal{H}\rightarrow[-cT,cT]$ and $Q_{s,t}^{\pi_{\theta}}(h,x):\mathcal{H}\times\mathcal{X}\rightarrow[-cT,cT]$ be given as follows:}

\vspace{-0.50em}

\begin{align}
V_{s,t}^{\pi_{\theta}}(h)
&
\coloneqq
\mathbb{E}_{\tau\sim p_{\theta}}[
\smallsum_{u=t}^{T}f_{s}(h_{u},x_{u})
|h_{t}=h]
\\
Q_{s,t}^{\pi_{\theta}}(h,x)
&
\coloneqq
\mathbb{E}_{\tau\sim p_{\theta}}[
\smallsum_{u=t}^{T}f_{s}(h_{u},x_{u})
|h_{t}=h,x_{t}=x]
\end{align}

\vspace{-0.25em}

\dayum{
where the notation for both $V_{s,t}^{\pi_{\theta}}$ and $Q_{s,t}^{\pi_{\theta}}$ is explicit as to their dependence on the policy $\pi_{\theta}$ being followed, the source $s$ under consideration, and the time $t$---unlike in typical imitation learning, we operate in a non-stationary (and non-Markovian) setting. For Lemma \ref{thm:local}, we require an additional result:}

\begin{relemma}[restate=thmpdl,name=Expected Quality Difference]\label{thm:pdl}\upshape
$
\Delta\bar{F}_{s}(\theta)
=
T
\mathbb{E}\hspace{-3pt}_{\substack{h\sim\mu_{s}\\x\sim\pi_{s}(\cdot|h)}}\hspace{-5pt}
Q_{s,t}^{\pi_{\theta}}(h,x)
-
T
\mathbb{E}\hspace{-3pt}_{\substack{h\sim\mu_{s}\\x\sim\pi_{\theta}(\cdot|h)}}\hspace{-5pt}
Q_{s,t}^{\pi_{\theta}}(h,x)
$.
\end{relemma}

\vspace{-1.00em}

\textit{Proof}. From Definition \ref{def:classifier},

\vspace{-0.50em}

\begin{align}
\Delta\bar{F}_{s}(\theta)
&
=
\mathbb{E}_{\tau\sim p_{s}}\textstyle\sum_{t}
f_{s}(h_{t},x_{t})
-
\mathbb{E}_{\tau\sim p_{\theta}}\textstyle\sum_{t}
f_{s}(h_{t},x_{t})
\\
&
=
\mathbb{E}_{\tau\sim p_{s}}\textstyle\sum_{t}
(f_{s}(h_{t},x_{t})+V_{s,t}^{\pi_{\theta}}(h_{t})-V_{s,t}^{\pi_{\theta}}(h_{t}))
-
\mathbb{E}_{\tau\sim p_{\theta}}\textstyle\sum_{t}
f_{s}(h_{t},x_{t})
\label{eq:telebefore}\\
&
=
\mathbb{E}_{\tau\sim p_{s}}\textstyle\sum_{t}
(f_{s}(h_{t},x_{t})+V_{s,t+1}^{\pi_{\theta}}(h_{t+1})-V_{s,t}^{\pi_{\theta}}(h_{t}))
\label{eq:teleafter}\\
&
=
\mathbb{E}_{\tau\sim p_{s}}\textstyle\sum_{t}
(Q_{s,t}^{\pi_{\theta}}(h_{t},x_{t})-V_{s,t}^{\pi_{\theta}}(h_{t}))
\\
&
=
T\mathbb{E}_{h\sim\mu_{s},x\sim\pi_{s}(\cdot|h)}
(Q_{s,t}^{\pi_{\theta}}(h,x)-V_{s,t}^{\pi_{\theta}}(h))
\\
&
=
T
\mathbb{E}_{h\sim\mu_{s},x\sim\pi_{s}(\cdot|h)}
Q_{s,t}^{\pi_{\theta}}(h,x)
-
T
\mathbb{E}_{h\sim\mu_{s},x\sim\pi_{\theta}(\cdot|h)}
Q_{s,t}^{\pi_{\theta}}(h,x)
\end{align}

\vspace{-0.50em}

where (\ref{eq:telebefore}) to (\ref{eq:teleafter}) telescopes terms, and we use the fact $V_{s,T+1}^{\pi}(h)=0$. This derivation can be view- ed as a non-stationary, non-Markovian analogue of the ``performance difference'' result in \cite{kakade2002approximately}. \QED

\thmlocal*

\vspace{-1.00em}

\textit{Proof}. From Lemma \ref{thm:pdl},

\vspace{-0.50em}

\begin{align}
\Delta\bar{F}_{s}(\theta)
&
=
T\mathbb{E}_{h\sim\mu_{s},x\sim\pi_{s}(\cdot|h)}
Q_{s,t}^{\pi_{\theta}}(h,x)
-
T\mathbb{E}_{h\sim\mu_{s},x\sim\pi_{\theta}(\cdot|h)}
Q_{s,t}^{\pi_{\theta}}(h)
\\
&
=
T
\mathbb{E}_{h\sim\mu_{s}}
[
\mathbb{E}_{x\sim\pi_{s}(\cdot|h)}
Q_{s,t}^{\pi_{\theta}}(h,x)
-
\mathbb{E}_{x\sim\pi_{\theta}(\cdot|h)}
Q_{s,t}^{\pi_{\theta}}(h,x)
]
\\
&
\leq
\text{max}_{Q\in[-cT,cT]^{\mathcal{H}\times\mathcal{X}}}
T
\mathbb{E}_{h\sim\mu_{s}}
[
\mathbb{E}_{x\sim\pi_{s}(\cdot|h)}
Q(h,x)
-
\mathbb{E}_{x\sim\pi_{\theta}(\cdot|h)}
Q(h,x)
]
\\
&
\leq
\text{max}_{f\in\mathbb{R}^{\mathcal{H}\times\mathcal{X}}}
T
\mathbb{E}_{h\sim\mu_{s}}
[
\mathbb{E}_{x\sim\pi_{s}(\cdot|h)}
Tf(h,x)
-
\mathbb{E}_{x\sim\pi_{\theta}(\cdot|h)}
Tf(h,x)
]
\\
&
\leq
T^{2}\epsilon
\end{align}

\vspace{-0.50em}

\dayum{
where the final inequality applies the assumption from the lemma.
Note that this is similar in spirit to various results for error accumulation in imitation learning through behavioral cloning. The most well-known one is \cite{ross2010efficient}, where a quadratic bound is given with respect to the \textit{probability} that the learned policy makes a small mistake. Another well-known one is in \cite{agarwal2019reinforcement}, where the bound is given with respect to \textit{sample complexity}. Here, in order to motivate our perspective from the notion of expected quality difference, our bound is given with respect to the \textit{moment-matching} discrepancy,
and can be interpreted as a non-Markovian variant of the ``off-policy upper bound'' result in \cite{swamy2021moments}. \QED}

\thmglobal*

\vspace{-1.00em}

\textit{Proof}. From Definition \ref{def:difference},

\vspace{-0.50em}

\begin{align}
\Delta\bar{F}_{s}(\theta)
&
=
\mathbb{E}_{\tau\sim p_{s}}\textstyle\sum_{t}
f_{s}(h_{t},x_{t})
-
\mathbb{E}_{\tau\sim p_{\theta}}\textstyle\sum_{t}
f_{s}(h_{t},x_{t})
\\
&
=
T\mathbb{E}_{h\sim\mu_{s},\pi_{s}(\cdot|h)}
f_{s}(h,x)
-
T\mathbb{E}_{h\sim\mu_{\theta},\pi_{\theta}(\cdot|h)}
f_{s}(h,x)
\\
&
\leq
\text{max}_{f\in[-c,c]^{\mathcal{H}\times\mathcal{X}}}
(
T\mathbb{E}_{h\sim\mu_{s},x\sim\pi_{s}(\cdot|h)}
f(h,x)
-
T\mathbb{E}_{h\sim\mu_{\theta},x\sim\pi_{\theta}(\cdot|h)}
f(h,x)
)
\\
&
\leq
\text{max}_{f\in\mathbb{R}^{\mathcal{H}\times\mathcal{X}}}
(
T\mathbb{E}_{h\sim\mu_{s},x\sim\pi_{s}(\cdot|h)}
f(h,x)
-
T\mathbb{E}_{h\sim\mu_{\theta},x\sim\pi_{\theta}(\cdot|h)}
f(h,x)
)
\\
&
\leq
T\epsilon
\end{align}

\vspace{-0.50em}

\dayum{
where the final inequality applies the assumption from the lemma.
Note that this is similar in spirit to various results for error accumulation in imitation learning through distribution matching. For instance, \cite{xu2021error} shows a bound in terms of \textit{divergences} in occupancy measures, while \cite{agarwal2019reinforcement} shows a bound in terms of \textit{sample complexity}. Here, in order to motivate our perspective from the notion of expected quality difference, our bound is given with respect to the \textit{moment-matching} discrepancy,
and can similarly be interpreted as a non-Markovian variant of the ``reward upper bound'' in \cite{swamy2021moments}. \QED}

\dayum{
For Propositions \ref{thm:opt} and \ref{thm:con}, we use the fact that training the structured classifier (Definition \ref{def:classifier}) using the energy loss (Equation \ref{eq:nceenergy}) amounts to a specific form of (sequence-wise) noise-contrastive estimation, and where the ``noise'' $p_{\theta}$ employed happens to be adaptively trained via the policy loss (Equation \ref{eq:ncepolicy}):}

\thmopt*

\dayum{
\textit{Proof}. Briefly, a noise-contrastive estimator \cite{gutmann2010noise} operates as follows: Suppose we have some data $y\in\mathcal{Y}$ distributed as $p_{\text{data}}(y)$. Consider that we wish to learn a model distribution $p_{\text{model}}$, as follows:}

\vspace{-0.25em}

\begin{equation}
p_{\text{model}}(y;a,b)
\coloneqq
\tilde{p}_{\text{model}}(y;a)\exp(b)
\end{equation}

\vspace{-0.25em}

parameterized by $a$ and $b$, where we emphasize that the model is not necessarily normalized as $b$ is simply a learnable parameter. Also denote any noise distribution that can be sampled and evaluated:

\vspace{-0.25em}

\begin{equation}
p_{\text{noise}}(y;c)
\end{equation}

\vspace{-0.25em}

parameterized by $c$. Now, define a classifier $d(\cdot\pix;a,b,c)$ as follows, which we shall train to discriminate between $p_{\text{data}}$ and $p_{\text{noise}}$---that is, given some $y$, to represent the (posterior) probability that it is real:

\vspace{-0.25em}

\begin{equation}
d(y;a,b,c)\coloneqq\sigma(\log p_{\text{model}}(y;a,b)-\log p_{\text{noise}}(y;c))
\end{equation}

\vspace{-0.25em}

where $\sigma$ indicates the usual sigmoid function, i.e. $\sigma(u)\coloneqq1/(1+\exp(-u))$ for any $u\in\mathbb{R}$. The noise contrastive estimator maximizes the likelihood of the parameters $a,b$ in $d$ given $p_{\text{data}}$ and $p_{\text{noise}}$:

\vspace{-0.25em}

\begin{equation}
\mathcal{L}_{\text{class}}(a,b;c)
\coloneqq
-
\mathbb{E}_{y\sim p_{\text{data}}}
\log d(y;a,b,c)
-
\mathbb{E}_{y\sim p_{\text{noise}}}
\log\big(1
-
d(y;a,b,c)\big)
\end{equation}

\vspace{-0.25em}

In this optimization problem, a basic result is that $\mathcal{L}_{\text{class}}$ attains a minimum at $\log p_{\text{model}}=\log p_{\text{data}}$ and that there are no other minima if $p_{\text{noise}}$ is chosen such that $p_{\text{data}}(y)>0\Rightarrow p_{\text{noise}}(y)>0$ holds: see the ``nonparametric estimation'' result in \cite{gutmann2012noise}. Now, let us consider the following correspondence:

\vspace{-0.25em}

\begin{equation}
\big(
\mathcal{Y}
,
p_{\text{data}}
,
\tilde{p}_{\text{model}}(\pix\cdot\pix;a)
,
b
,
p_{\text{noise}}(\pix\cdot\pix;c)
\big)
\coloneqq
\big(
\mathcal{T}
,
p_{s}
,
\tilde{p}_{\phi}
,
-\log Z_{\phi}
,
p_{\theta}
\big)
\end{equation}

\vspace{-0.25em}

In other words, let the underlying space be that of trajectories $\mathcal{Y}\coloneqq\mathcal{T}$; let the data distribution be $p_{\text{data}}\coloneqq p_{s}$; let the model distribution be given by the un-normalized energy model $\tilde{p}_{\text{model}}(\pix\cdot\pix;a)\coloneqq\tilde{p}_{\phi}$ and partition function $b=-\log Z_{\phi}$; and let the noise distribution be given by rollouts of the policy, $p_{\text{noise}}(\pix\cdot\pix;c)\coloneqq p_{\theta}$. Then it is easy to see that the classifier and its loss function correspond as follows:

\vspace{-0.25em}

\begin{equation}
\begin{aligned}
d_{\theta,\phi}(\pix\cdot\pix)&=d(\pix\cdot\pix\pix;a,b,c)
\\
\mathcal{L}_{\text{energy}}(\phi;\theta)&=\mathcal{L}_{\text{class}}(a,b;c)
\end{aligned}
\end{equation}

\vspace{-0.25em}

But then the optimality result above directly maps to the statement that $\mathcal{L}_{\text{energy}}$ is globally minimized at $F_{\phi}(\cdot)-\log Z_{\phi}=\log p_{s}(\cdot)$ assuming that the positivity condition $p_{s}(\tau)>0\Rightarrow p_{\theta}(\tau)>0$ holds. Technicality: Note that here $\tilde{p}_{\phi}$ is constrained as $F_{\phi}(\tau)\coloneqq\sum_{t}f_{\phi}(h_{t},x_{t})$ instead of being arbitrarily parameterizable, but this does not affect realizability as we assumed that $F_{s}(\tau)\coloneqq\sum_{t}f_{s}(h_{t},x_{t})$. \QED

\thmcon*

\vspace{-0.25em}

\textit{Proof}. Continuing the exposition above, let $\mathcal{L}_{\text{class}}^{M}(a,b;c)$ indicate the finite-data approximation of $\mathcal{L}_{\text{class}}(a,b;c)$---that is, by using $M$ samples to approximate the true expectations over $p_{\text{data}}$ and $p_{\text{noise}}$:

\vspace{-0.25em}

\begin{equation}
\mathcal{L}_{\text{class}}^{M}(a,b;c)
\coloneqq
-
\tfrac{1}{M}\textstyle\sum_{m=1}^{M}
\log d(y_{\text{data}}^{(m)};a,b,c)
-
\tfrac{1}{M}\textstyle\sum_{m=1}^{M}
\log\big(1
-
d(y_{\text{noise}}^{(m)};a,b,c)\big)
\end{equation}

\vspace{-0.25em}

where the samples are drawn as \smash{$\raisebox{1pt}{$y_{\text{data}}^{(m)}$}\sim p_{\text{data}}$} and \smash{$\raisebox{1pt}{$y_{\text{noise}}^{(m)}$}\sim p_{\text{noise}}$}. Consider the following conditions:

\begin{enumerate}[leftmargin=1.25em]
\itemsep0pt
\item Positivity: \smash{$p_{\text{data}}(y)>0\Rightarrow p_{\text{noise}}(y)>0$};
\item Uniform convergence: \smash{$\sup_{a,b}|\mathcal{L}_{\text{class}}^{M}(a,b;c)-\mathcal{L}_{\text{class}}(a,b;c)|\overset{_p}{\longrightarrow}0$}; and
\item The following matrix is full-rank: $\mathcal{I}\coloneqq$ \smash{$\int g(y)g(y)$}\smash{\raisebox{-1.5pt}{$^{\top}$}}\smash{$p_{\text{data}}(y)p_{\text{noise}}(y)/(p_{\text{data}}(y)$\pix$+$\pix$p_{\text{noise}}(y))dy$}, where \smash{$g(y)\coloneqq\nabla_{(a,b)}\log p_{\text{model}}(y;a,b)|_{(a^{*},b^{*})}$} and $a^{*},b^{*}$ denote the optimal values of the model.
\end{enumerate}

Note that (1) is same as before, and (2) and (3) are analogous to standard assumptions in maximum likelihood estimation. Let \smash{$\hat{a}_{M}^{*},\hat{b}_{M}^{*}$} denote the minimizers for \smash{$\mathcal{L}_{\text{class}}^{M}(a,b;c)$}. Under the preceding conditions, another basic result is that \smash{$(\hat{a}_{M}^{*},\hat{b}_{M}^{*})$} converges in probability to \smash{$(a^{*},b^{*})$} as $M$ grows: see the ``consistency'' result in \cite{gutmann2012noise}. But continuing the correspondence from before, it is easy to see that

\vspace{-0.25em}

\begin{equation}
\mathcal{L}_{\text{energy}}^{M}(\phi;\theta)=\mathcal{L}_{\text{class}}^{M}(a,b;c)
\end{equation}

\vspace{-0.25em}

where we similarly define $\mathcal{L}_{\text{energy}}^{M}(\phi;\theta)$ to be the finite-data approximation of $\mathcal{L}_{\text{energy}}(\phi;\theta)$---that is, by using $M$ samples to approximate the true expectations over $p_{s}$ and $p_{\theta}$, and \smash{$y_{s}^{_{(m)}}\hspace{-5pt}\sim p_{s}$} and \smash{$y_{\theta}^{_{(m)}}\hspace{-5pt}\sim p_{\theta}$}:

\vspace{-0.25em}

\begin{equation}
\mathcal{L}_{\text{energy}}^{M}(\phi;\theta)
\coloneqq
-
\tfrac{1}{M}\textstyle\sum_{m=1}^{M}
\log d_{\theta,\phi}(\tau_{s}^{(m)})
-
\tfrac{1}{M}\textstyle\sum_{m=1}^{M}
\log\big(1
-
d_{\theta,\phi}(\tau_{\theta}^{(m)})\big)
\end{equation}

\vspace{-0.25em}

which directly maps the above convergence result to the statement that as $M$ increases \smash{$\hat{\phi}_{M}^{*}\overset{_p}{\rightarrow}\phi^{*}$}. \QED

\thmsce*

\newcommand{\spacer}{\vphantom{\mfrac{\nabla_{\phi}p_{\phi}(\tau)}{\log(p_{\phi}(\tau)+p_{\theta_{k}^{*}}(\tau))}}}

\vspace{-1.00em}

\textit{Proof}. From Equation \ref{eq:nceenergy},

\vspace{-0.50em}

\begin{align}
\nabla_{\phi}
\mathcal{L}&_{\text{energy}}(\phi;\theta_{k}^{*})
=\spacer
\nabla_{\phi}\big(
-
\mathbb{E}_{\tau\sim p_{s}}
\log d_{\theta_{k}^{*},\phi}(\tau)
-
\mathbb{E}_{\tau\sim p_{\theta_{k}^{*}}}
\log\big(1
-
d_{\theta_{k}^{*},\phi}(\tau)\big)
\big)
\\
&
=\spacer
\nabla_{\phi}\big(
-
\mathbb{E}_{\tau\sim p_{s}}
\log\mfrac{p_{\phi}(\tau)}{p_{\phi}(\tau)+p_{\theta_{k}^{*}}(\tau)}
-
\mathbb{E}_{\tau\sim p_{\theta_{k}^{*}}}
\log\mfrac{p_{\theta_{k}^{*}}(\tau)}{p_{\phi}(\tau)+p_{\theta_{k}^{*}}(\tau)}
\big)
\\
&
=\spacer
-
\mathbb{E}_{\tau\sim p_{s}}
\nabla_{\phi}(
\log p_{\phi}(\tau)-\log(p_{\phi}(\tau)+p_{\theta_{k}^{*}}(\tau))
)
+
\mathbb{E}_{\tau\sim p_{\theta_{k}^{*}}}
\nabla_{\phi}
\log(p_{\phi}(\tau)+p_{\theta_{k}^{*}}(\tau))
\\
&
=\spacer
-
\mathbb{E}_{\tau\sim p_{s}}\big[
\nabla_{\phi}
\log p_{\phi}(\tau)-\mfrac{\nabla_{\phi}p_{\phi}(\tau)}{p_{\phi}(\tau)+p_{\theta_{k}^{*}}(\tau)}
\big]
+
\mathbb{E}_{\tau\sim p_{\theta_{k}^{*}}}
\mfrac{\nabla_{\phi}p_{\phi}(\tau)}{p_{\phi}(\tau)+p_{\theta_{k}^{*}}(\tau)}
\\
&
=\spacer
-
\mathbb{E}_{\tau\sim p_{s}}\big[
\nabla_{\phi}
\log p_{\phi}(\tau)-\mfrac{p_{\phi}(\tau)\nabla_{\phi}\log p_{\phi}(\tau)}{p_{\phi}(\tau)+p_{\theta_{k}^{*}}(\tau)}
\big]
+
\mathbb{E}_{\tau\sim p_{\theta_{k}^{*}}}
\mfrac{p_{\phi}(\tau)\nabla_{\phi}\log p_{\phi}(\tau)}{p_{\phi}(\tau)+p_{\theta_{k}^{*}}(\tau)}
\\
&
=\spacer
-
\mathbb{E}_{\tau\sim p_{s}}\big[
\nabla_{\phi}
\log p_{\phi}(\tau)-\tfrac{1}{2}\nabla_{\phi}\log p_{\phi}(\tau)
\big]
+
\tfrac{1}{2}
\mathbb{E}_{\tau\sim p_{\theta_{k}^{*}}}
\nabla_{\phi}\log p_{\phi}(\tau)
\\
&
=
\tfrac{1}{2}
\mathbb{E}_{\tau\sim p_{\theta_{k}^{*}}}
\nabla_{\phi}\log p_{\phi}(\tau)
-
\tfrac{1}{2}
\mathbb{E}_{\tau\sim p_{s}}
\nabla_{\phi}\log p_{\phi}(\tau)
\\
&
=
\tfrac{1}{2}
\mathbb{E}_{\tau\sim p_{\theta_{k}^{*}}}
\nabla_{\phi}(\log\tilde{p}_{\phi}(\tau)
-
\log Z_{\phi})
-
\tfrac{1}{2}
\mathbb{E}_{\tau\sim p_{s}}
\nabla_{\phi}(\log\tilde{p}_{\phi}(\tau)
-
\log Z_{\phi}
)
\\
&
=
\tfrac{1}{2}\big(
\mathbb{E}_{\tau\sim p_{\theta_{k}^{*}}}
\nabla_{\phi}\textstyle\sum_{t}f_{\phi}(h_{t},x_{t})
-
\mathbb{E}_{\tau\sim p_{s}}
\nabla_{\phi}\textstyle\sum_{t}f_{\phi}(h_{t},x_{t})
\big)
\\
&
=
\tfrac{1}{2}\big(
T\mathbb{E}_{h\sim\mu_{\theta_{k}^{*}},x\sim\pi_{\theta_{k}^{*}}(\cdot|h)}
\nabla_{\phi}\textstyle f_{\phi}(h,x)
-
T\mathbb{E}_{h\sim\mu_{s},x\sim\pi_{s}(\cdot|h)}
\nabla_{\phi}\textstyle f_{\phi}(h,x)
\big)
\\
&
=
-\tfrac{T}{2}\nabla_{\phi}\mathcal{L}(\theta_{k}^{*},\phi)
\end{align}

\vspace{-0.25em}

\dayum{
where the fourth and fifth lines repeatedly use the identity $\nabla_{z}p_{z}\equiv p_{z}\nabla_{z}\log p_{z}$ for any $p_{z}$ parameterized by $z$, and the sixth line uses the fact that the current value of $\theta$ (i.e. $\theta_{k}^{*}$) is the minimizer for $\mathcal{L}_{\text{policy}}(\theta;\phi)$ at the current value of $\phi$ (i.e. $\phi_{k}$), hence it must be the case that $p_{\theta}=p_{\phi}$~at~those~values. Note that this assumes that $p_{\phi}$ is normalized; in practice this will be approximately true, for instance if we pre-train $\phi$ beforehand, using a fixed $\pi_{\theta}$ pre-trained by maximum likelihood (see Appendix \ref{app:b}).}

\dayum{
In the more general case of any arbitrary un-normalized $p_{\phi}$, we only know \smash{$\raisebox{1pt}{$p_{\theta_{k}^{*}}$}=\tfrac{1}{K_{\phi}}p_{\phi}$} for some constant $K_{\phi}$; then we recover a generalized ``weighted'' version of Equation \ref{eq:vebmenergy}. From the fifth line above,}

\vspace{-0.50em}

\begin{align}
\nabla_{\phi}
\mathcal{L}&_{\text{energy}}(\phi;\theta_{k}^{*})
=
-
\mathbb{E}_{\tau\sim p_{s}}\big[
\nabla_{\phi}
\log p_{\phi}(\tau)-\mfrac{p_{\phi}(\tau)\nabla_{\phi}\log p_{\phi}(\tau)}{p_{\phi}(\tau)+p_{\theta_{k}^{*}}(\tau)}
\big]
+
\mathbb{E}_{\tau\sim p_{\theta_{k}^{*}}}
\mfrac{p_{\phi}(\tau)\nabla_{\phi}\log p_{\phi}(\tau)}{p_{\phi}(\tau)+p_{\theta_{k}^{*}}(\tau)}
\\
&
=
-
\mathbb{E}_{\tau\sim p_{s}}\big[
\nabla_{\phi}
\log p_{\phi}(\tau)-\tfrac{K_{\phi}}{K_{\phi}+1}\nabla_{\phi}\log p_{\phi}(\tau)
\big]
+
\tfrac{K_{\phi}}{K_{\phi}+1}
\mathbb{E}_{\tau\sim p_{\theta_{k}^{*}}}
\nabla_{\phi}\log p_{\phi}(\tau)
\\
&
=
\tfrac{K_{\phi}}{K_{\phi}+1}
\mathbb{E}_{\tau\sim p_{\theta_{k}^{*}}}
\nabla_{\phi}\log p_{\phi}(\tau)
-
\tfrac{1}{K_{\phi}+1}
\mathbb{E}_{\tau\sim p_{s}}
\nabla_{\phi}\log p_{\phi}(\tau)
\\
&
=
\tfrac{K_{\phi}}{K_{\phi}+1}
\mathbb{E}_{\tau\sim p_{\theta_{k}^{*}}}
\nabla_{\phi}(\log\tilde{p}_{\phi}(\tau)
-
\log Z_{\phi})
-
\tfrac{1}{K_{\phi}+1}
\mathbb{E}_{\tau\sim p_{s}}
\nabla_{\phi}(\log\tilde{p}_{\phi}(\tau)
-
\log Z_{\phi}
)
\\
&
=
\tfrac{K_{\phi}}{K_{\phi}+1}
\mathbb{E}_{\tau\sim p_{\theta_{k}^{*}}}
\nabla_{\phi}\textstyle\sum_{t}f_{\phi}(h_{t},x_{t})
-
\tfrac{1}{K_{\phi}+1}
\mathbb{E}_{\tau\sim p_{s}}
\nabla_{\phi}\textstyle\sum_{t}f_{\phi}(h_{t},x_{t})
\\
&
=
\tfrac{TK_{\phi}}{K_{\phi}+1}
\mathbb{E}_{h\sim\mu_{\theta_{k}^{*}},x\sim\pi_{\theta_{k}^{*}}(\cdot|h)}
\nabla_{\phi}\textstyle f_{\phi}(h,x)
-
\tfrac{T}{K_{\phi}+1}
\mathbb{E}_{h\sim\mu_{s},x\sim\pi_{s}(\cdot|h)}
\nabla_{\phi}\textstyle f_{\phi}(h,x)
\end{align}

\vspace{-0.50em}

This ``weighting'' is intuitive: If $p_{\phi}$ is un-normalized such that $K_{\phi}>1$, the energy loss automatically places higher weights on negative samples $h\sim\raisebox{1pt}{$\mu_{\theta_{k}^{*}}$},x\sim\raisebox{1pt}{$\pi_{\theta_{k}^{*}}$}(\cdot|h)$ to bring it down; conversely, if $p_{\phi}$ is un-normalized such that $K_{\phi}<1$, the energy loss places higher weights on positive samples $h\sim\mu_{s},x\sim\pi_{s}(\cdot|h)$ to bring it up. (If $p_{\phi}$ is normalized, then $K_{\phi}=1$ and the weights are equal). \QED

\dayum{
\textbf{Note on Duality}~
Lemmas 1 and 2 are for building intuition, and we are \textit{not} implying that Equations \ref{eq:local2} and \ref{eq:global2} are direct generalizations of the duality between maximum likelihood and maximum entropy. To be clear, we shall explain Equation \ref{eq:global2}; Equation \ref{eq:local2} is similar. Consider first the case of finite $\mathcal{X}$ (therefore finite $\mathcal{T}$). In the usual linear case, given basis functions $T(\tau)$, the optimization problem is:}

\vspace{-0.25em}

\begin{equation}
\text{arg\pix min}_{\theta}E_{\tau\sim p_{\theta}}\log p_{\theta}(\tau)\quad\text{s.t.}\quad E_{\tau\sim p_{s}}T(\tau)=E_{\tau\sim p_{\theta}}T(\tau)
\end{equation}
Internalizing the constraint, we may write $\text{arg\pix min}_{\theta}(E_{\tau\sim p_{\theta}}\log p_{\theta}(\tau)+\max_{F}(\langle F,E_{\tau\sim p_{s}}T(\tau)\rangle-\langle F,E_{\tau\sim p_{\theta}}T(\tau)\rangle))$. To generalize to the nonlinear case, let us now specifically define the feature vector $T(\tau)$ to be the indicator function (i.e. a finite-length vector, each zero-one entry of which corresponds to each element in $\mathcal{T}$). Then note that $\langle F,E_{\tau\sim p}T(\tau)\rangle=\langle F,p\rangle=E_{\tau\sim p}F(\tau)$, therefore:
\begin{equation}
\text{arg\pix min}_{\theta}(E_{\tau\sim p_{\theta}}\log p_{\theta}(\tau)+\max_{F}(E_{\tau\sim p_{s}}F(\tau)-E_{\tau\sim p_{\theta}}F(\tau)))
\end{equation}
\dayum{%
Finally, to arrive at Equation \ref{eq:global2} we use the fact that $E_{\tau\sim p}\sum_{t}f(h_{t},x_{t})=T E_{h\sim\mu,x\sim\pi(\cdot|h)}f(h,x)$, as already noted. All in all, this is a ``generalization'' from the case of linearity in \textit{known} basis functions, to the case of \textit{unknown} basis functions---where we ``linearize'' the expression using indicator functions. But we cannot use the same logic to claim the same for \textit{infinite} $\mathcal{X}$ (which is the setting we operate in).}


\section{Details on Algorithm}\label{app:b}

\dayum{
\textbf{Policy Optimization}~
Recall the policy update (Equation \ref{eq:ncepolicy}); this corresponds to entropy-regularized reinforcement learning using $f_{\phi}(h,x)$ as transition-wise reward function. Here we give a brief review of entropy-regularized reinforcement learning \cite{fox2016taming,haarnoja2017reinforcement,shi2019soft} in our context, as well as the practical method we employ (i.e. soft actor-critic). First, we introduce some standard notation. At any state $h$, define the (soft) ``value function'' to be the (forward-looking) expected sum of future rewards $f_{\phi}(h,x)$ as well as entropies $H(\pi(\cdot|h))$. Specifically, let \smash{$V_{\phi}^{\pi_{\theta}}(h)$} and \smash{$Q_{\phi}^{\pi_{\theta}}(h,x)$} be given as follows (we omit explicit notation for $t$, as any influence of time is implicit through dependence on variable-length histories):}

\vspace{-0.50em}

\begin{align}
V_{\phi}^{\pi_{\theta}}(h)
&
\coloneqq
\mathbb{E}_{\tau\sim p_{\theta}}[
\smallsum_{u=t}^{T}f_{\phi}(h_{u},x_{u})
+
H(\pi_{\theta}(\cdot|h_{u}))
|h_{t}=h]
\\
Q_{\phi}^{\pi_{\theta}}(h,x)
&
\coloneqq
f_{\phi}(h,x)
+
\mathbb{E}_{\tau\sim p_{\theta}}[
\smallsum_{u=t+1}^{T}f_{\phi}(h_{u},x_{u})
+
H(\pi_{\theta}(\cdot|h_{u}))
|h_{t}=h,x_{t}=x]
\end{align}

\vspace{-0.50em}

Let \smash{$\pi_{\theta^{*}}$} denote the optimal policy (i.e. that minimizes loss $\mathcal{L}_{\text{policy}}$), and \smash{$Q_{\phi}^{\pi_{\theta^{*}}}$} its corresponding value function. An elementary result is that the optimal policy assigns probabilities to $x$ proportional to the exponentiated expected returns of energy and entropy terms of all trajectories that begin with $(h,x)$:

\vspace{-0.75em}

\begin{equation}
\pi_{\theta^{*}}(x|h)=\mfrac{\exp(Q_{\phi}^{\pi_{\theta^{*}}}(h,x))}{\int_{\mathcal{X}}\exp(Q_{\phi}^{\pi_{\theta^{*}}}(h,x))dx}
\end{equation}

\vspace{-0.75em}

Now, for any transition policy $\pi_{\theta}$ (i.e. not necessarily optimal with respect to $f_{\phi}$), the value function \smash{$Q_{\phi}^{\pi_{\theta}}$} is the unique fixed point of the following (soft) Bellman backup operator \smash{$\mathbb{B}_{\phi}^{\pi_{\theta}}$\pix$:$\pix$\mathbb{R}^{\mathcal{H}\times\mathcal{X}}$$\rightarrow$\pix$\mathbb{R}^{\mathcal{H}\times\mathcal{X}}$}:

\vspace{-0.50em}

\begin{equation}
(\mathbb{B}_{\phi}^{\pi_{\theta}}
Q
)
(h,x)
\coloneqq
f_{\phi}(h,x)
+
\mathbb{E}_{x'\sim\pi_{\theta}(\cdot|h')}[Q(h',x')-\log\pi_{\theta}(x'|h')]
\end{equation}

\vspace{-0.50em}

and hence---in theory---\smash{$Q_{\phi}^{\pi_{\theta}}$} may be computed iteratively by repeatedly applying the operator \smash{$\mathbb{B}_{\phi}^{\pi_{\theta}}$} starting from any function $Q\in\mathbb{R}^{\mathcal{H}\times\mathcal{X}}$; this is referred to as the (soft) ``policy evaluation'' procedure.

Using \smash{$Q_{\phi}^{\pi_{\theta}}$}, we may then perform (soft) ``policy improvement'' to update the policy $\pi_{\theta}$ towards the exponential of its value function, and is guaranteed to result in an improved policy (in terms of \smash{$Q_{\phi}^{\pi_{\theta}}$}):

\vspace{-0.75em}

\begin{equation}
\theta'
\leftarrow
\text{arg\pix min}_{\theta}~
D_{\text{KL}}\Big(\pi_{\theta'}(\cdot|h)\big\|\mfrac{\exp(Q_{\phi}^{\pi_{\theta}}(h,\cdot))}{\int_{\mathcal{X}}\exp(Q_{\phi}^{\pi_{\theta}}(h,x))dx}\Big)
\end{equation}

\vspace{-0.75em}

\dayum{
for all $h\in\mathcal{H}$. In theory, then, finding the optimal policy can be approached by repeatedly applying the above policy evaluation and policy improvement steps starting from any initial policy $\pi_{\theta}$; this is referred to as (soft) ``policy iteration''. However, in large continuous domains (such as $\mathcal{H}\times\mathcal{X}$) doing this exactly is impossible, so we need to rely on function approximation for representing value functions.}

\dayum{
\textbf{Practical Algorithm}~
Precisely, the soft actor-critic approach is to introduce a function approximator to represent the value function (i.e. the ``critic'') parameterized by $\psi$, in addition to the policy itself (i.e. the ``actor'') parameterized by $\theta$, and to alternate between optimizing both with stochastic gradient descent \cite{haarnoja2018soft}. Specifically, the actor performs soft policy improvement steps as before, but now using $Q_{\psi}$:}

\vspace{-0.25em}

\begin{equation}\label{eq:finalactor}
\mathcal{L}_{\text{actor}}(\theta;\phi,\psi)
\coloneqq
\mathbb{E}_{h\sim\mathcal{B}}~
\mathbb{E}_{x\sim\pi_{\theta}(\cdot|h)}
[\log\pi_{\theta}(x|h)-Q_{\psi}(h,x)]
\end{equation}

\vspace{-0.25em}

where $\mathcal{B}$ is a replay buffer of samples generated by $\pi_{\theta}$ (that is, instead of enumerating all $h\in\mathcal{H}$, we are relying on $h\sim\mathcal{B}$). Note that the normalizing constant is dropped as it does not contribute to the gradient. The critic is trained to represent the value function by minimizing squared residual errors:

\vspace{-0.25em}

\begin{equation}\label{eq:finalcritic}
\mathcal{L}_{\text{critic}}(\psi;\phi)
\coloneqq
\mathbb{E}_{h,x\sim\mathcal{B}}
(
Q_{\psi}(h,x)-Q_{\psi}^{\text{target}}(h,x)
)^{2}
\end{equation}

\vspace{-0.75em}

with (bootstrapped) targets:

\vspace{-0.50em}

\begin{equation}
Q_{\psi}^{\text{target}}(h,x)
\coloneqq
f_{\phi}(h,x)
+
\mathbb{E}_{x'\sim\pi_{\theta}(\cdot|h')}[
Q_{\psi}(h',x')-\log\pi_{\theta}(x'|h')
]
\end{equation}

\vspace{-0.25em}

\dayum{
Together, this provides a way to minimize the policy loss $\mathcal{L}_{\text{policy}}$ (Equation \ref{eq:ncepolicy}). The complete Time- GCI algorithm simply alternates between this and minimizing the energy loss $\mathcal{L}_{\text{energy}}$ (Equation \ref{eq:nceenergy}):}

\vspace{-0.25em}

\begin{equation}\label{eq:finalenergy}
\mathcal{L}_{\text{energy}}(\phi;\theta)
\coloneqq
\text{$-$}
\mathbb{E}_{\tau\sim p_{s}}
\log d_{\theta,\phi}(\tau)
\text{$-$}
\mathbb{E}_{\tau\sim p_{\theta}}
\log\big(1
\text{$-$}
d_{\theta,\phi}(\tau)\big)
\end{equation}

\vspace{-0.25em}

\dayum{
Hence in Algorithm \ref{alg:timegci}, gradient updates for the energy, policy, and critic are interleaved with policy rollouts. Note that several standard approximations are being used. First, the replay buffer provides samples $h\sim\mathcal{B}$ for optimizing the policy (in both actor and critic updates), instead of covering the entire space $\mathcal{H}$ (which is uncountable). Second, in the energy loss negative samples $\tau\sim\mathcal{B}$ are used in lieu of sampling fresh from $p_{\theta}$ at every iteration (this is known to give the benefit of providing more diverse negative samples). Finally, also per usual samples from the dataset $\tau\sim\mathcal{D}$ is used in lieu of $p_{s}$.}

\dayum{
\textbf{Practical Considerations}~
First, in practice we must use a \textit{vector representation} of histories $h\in\mathcal{H}$; here we use RNNs to encode histories into fixed-length vectors, which can then be treated as regular ``states'' in continuous space. Like our choice of policy optimization, this is also an arbitrary design choice---we could just as conceivably have used e.g. temporal convolutions, attention mechanisms, etc.}

\dayum{
Second, \textit{interleaving} multiple gradient updates of different networks requires some care: In soft actor-critic itself, policy updates have to be sufficiently small, and/or critic updates have to be sufficiently frequent, to prevent divergence. The situation is analogous when interleaving this with energy gradient updates as well: Both actor and energy updates have to be sufficiently small, and/or critic updates have to be sufficiently frequent. That said, the energy updates are indeed decoupled from the policy updates: Regardless of how quickly/slowly the policy is learning, the energy can learn on their negative samples. In practice, we perform multiple critic updates for every update of the policy and energy functions.}

\dayum{
Finally, note that in large continuous domains such as $\mathcal{H}\times\mathcal{X}$ it is necessary to \textit{pre-train} the networks beforehand such that optimization of the complete algorithm actually converges: On the one hand, the policy side requires a sufficiently good energy signal to actually make progress, and on the other hand, the energy side requires a sufficiently good policy providing challenging enough negative samples to actually make progress. Pre-training networks separately is standard in actor-critic methods (see for instance \cite{bahdanau2017actor}); here we take a similar approach but with the addition of the energy update step as well:}

\begin{enumerate}[leftmargin=1.25em]
\itemsep0pt
\item Policy-only: $\pi_{\theta}$ is pre-trained using maximum likelihood;
\item Energy-only: $f_{\phi}$ is pre-trained using $\mathcal{L}_{\text{energy}}(\phi;\theta)$, holding $\pi_{\theta}$ fixed;
\item Critic-only: $Q_{\psi}$ is pre-trained using $\mathcal{L}_{\text{critic}}(\psi;\phi)$, holding $\pi_{\theta}$, $f_{\phi}$ fixed; and finally,
\item \dayum{All: $f_{\phi}$, $\pi_{\theta}$, and $Q_{\psi}$ are trained on $\mathcal{L}_{\text{energy}}(\phi;\theta)$, $\mathcal{L}_{\text{actor}}(\theta;\phi,\psi)$, and $\mathcal{L}_{\text{critic}}(\psi;\phi)$ (cf. Algorithm \ref{alg:timegci}).}
\end{enumerate}

%
%
%
%
%
%

%
%
%

\section{Details on Experiments}\label{app:c}

\textbf{Benchmark Algorithms}~
Except where components are standardized (see below), we use the publicly available source code when constructing the benchmark algorithms; references are in the following:

\vspace{-0.45em}

\begin{itemize}[leftmargin=1.25em]\small
\itemsep-1.5pt
\item T-Forcing \cite{williams1989learning}: (straightforward MLE with ground-truth conditioning)
\item P-Forcing \cite{lamb2016professor}: \url{https://github.com/anirudh9119/LM_GANS}
\item C-RNN-GAN \cite{mogren2016c}: \url{https://github.com/olofmogren/c-rnn-gan}
\item COT-GAN \cite{xu2020cot}: \url{https://github.com/tianlinxu312/cot-gan}
\item RC-GAN \cite{esteban2017real}: \url{https://github.com/ratschlab/RGAN}
\item TimeGAN \cite{yoon2019time}: \url{https://github.com/jsyoon0823/TimeGAN}
\end{itemize}

\vspace{-0.40em}

\dayum{
\textbf{Dataset Sources}~
We use the original source code for preprocessing sines and UCI datasets from TimeGAN (\url{https://github.com/jsyoon0823/TimeGAN}). For MIMIC-III, we extract 52 clinical covariates including vital signs (e.g. respiratory rate, heart rate, O2 saturation) and lab tests (e.g. glucose, hemoglobin, white blood cell count) aggregated every hour during their ICU stay up to 24 hours.}

\vspace{-0.45em}

\begin{itemize}[leftmargin=1.25em]\small
\itemsep-1.5pt
\item Sines \cite{williams1989learning}: \url{https://github.com/jsyoon0823/TimeGAN}
\item Energy \cite{lamb2016professor}: {\url{archive.ics.uci.edu/ml/datasets/Appliances+energy+prediction}}
\item Gas \cite{mogren2016c}: {\small\url{archive.ics.uci.edu/ml/datasets/Gas+sensor+array+temperature+modulation}}
\item Metro \cite{xu2020cot}: {\url{archive.ics.uci.edu/ml/datasets/Metro+Interstate+Traffic+Volume}}
\item MIMIC-III \cite{esteban2017real}: \url{https://physionet.org/content/mimiciii/1.4/}
\end{itemize}

\vspace{-0.40em}

\dayum{
\textbf{Encoder Networks}~
For fair comparison, analogous network components across all benchmarks share the same architecture where possible. In particular, all components taking $h_{t}$ as input require an encoder network to construct fixed-length vector representations of variable-length histories $(x_{1},...,x_{t})$. To do so, these components use an LSTM network with a hidden layer of size 32 to compute hidden states for representing $h$. These are not shared: separate components have their own encoder networks.}

\dayum{
\textbf{Task-Specific Networks}~
Then, for mapping from $h_{t}$ and/or $x_{t}$ to task-specific output variables, we use a fully-connected network with two hidden layers of size 32 and ELU activations. Where both $h_{t}$ and $x_{t}$ serve as inputs, their vectors are concatenated. For instance, the energy network for TimeGCI contains one such network for computing $f_{\phi}$ (as well as trainable parameter for $Z_{\phi}$). The same applies to the mapping from $h_{t}$ and/or $x_{t}$ to implicit generator outputs (as in C-RNN-GAN and RC-GAN), black-box discriminator output (as in P-Forcing, C-RNN-GAN, and RC-GAN), transition policies (as in T-Forcing and P-Forcing), and critic values (for TimeGCI). Note that TimeGAN and COT-GAN were designed with additional unique components/losses that operate as a unit (e.g. the embedding/ recovery networks for generating/discriminating within latent space); we use their original source code.}

\dayum{
\textbf{Replay Buffer}~
The replay buffer $\mathcal{B}$ has a fixed size; once filled, new samples stored replace the oldest still in the buffer. Sampling from the buffer operates as follows:  In updating the energy, we require $\tau\sim\mathcal{B}$ (that is, in lieu of $p_{\theta}$, cf. Equation \ref{eq:finalenergy}); this is done by randomly sampling a batch of trajectories from the replay buffer, without replacement. In policy the actor, we require $h\sim\mathcal{B}$ (cf. Equation \ref{eq:finalactor}); this is done by first randomly sampling a batch of trajectories from the replay buffer, and then randomly sampling a cutoff time $t$ to obtain a batch of subsequences $h_{t}$. Finally, in updating the critic, we require $h,x\sim\mathcal{B}$ (cf. Equation \ref{eq:finalcritic}); this is similarly done by first randomly sampling a batch of trajectories from the replay buffer, then randomly sampling a cutoff $t$ to yield a batch of $(h_{t},x_{t})$ pairs.}

\dayum{
\textbf{Hyperparameters}~
In all experiments, we use the following hyperparameters for TimeGCI and for benchmarks (wherever applicable): The replay buffer is of size $|\mathcal{B}|=10,000$. The hidden dimension of all encoder and task-specific networks is 32. The entropy regularization (in the actor/policy loss) is $\alpha=0.2$. The policy network is pre-trained for 2,000 steps, energy for 4,000, and critic for 20,000. The complete algorithm is trained for up to 50,000 steps with checkpointing and early stopping (triggerable every 1,000 iterations, if performance does not improve). The Adam optimizer is used for all losses, and batch size $M=64$. Learning rates: For energy networks $\lambda_{\text{energy}}=0.0001$, for policy networks $\lambda_{\text{policy}}=0.0001$ (same for implicit generator networks), for critic networks $\lambda_{\text{critic}}=0.001$, and for black-box discriminator networks $\lambda_{\text{discrim}}=0.001$. (Note that the critic/discriminator networks are updated more greedily). Per usual in soft actor-critic algorithms, we also employ a lagged target critic network (i.e. used for bootstrapping); this is updated using polyak averaging at a rate of $\tau=0.005$.}

\dayum{
\textbf{Performance Metrics}~
We use the original source code for computing the TSTR metric (i.e. Predictive Score), publicly available at: \url{https://github.com/jsyoon0823/TimeGAN}; this is straightforwardly modified to compute TSTR scores for horizons of lengths three (+3 Steps Ahead) and five (+5 Steps Ahead). Likewise, we use the original source code for computing the cross-correlation score ($x$-Corr. Score), this is also publicly available at: \url{https://github.com/tianlinxu312/cot-gan}.}

\section{Clarifying the Analogy}\label{app:d}

\dayum{This section clarifies the analogy of ``generation as imitation''. While the point of our investigation is to explicitly invite an analogy between synthetic time-series generation and imitation learning, they are different problems with different considerations. In particular, TimeGCI is not suitable for imitation learning per se, and we make no claim that it is (Appendix \ref{app:d1}). Conversely, while existing adversarial imitation learning methods may be naively applied to time-series generation, their optimization objectives are different, and we empirically verify they do not perform well (Appendix \ref{app:d2}).}

\subsection{Can TimeGCI be used for Imitation Learning?}\label{app:d1}

\dayum{
At its core, time-series generation is the problem of modeling a distribution of trajectories $p(\tau)$ faithfully, and is what TimeGCI does. In MDP parlance, in time-series generation the ``environment dynamics'' are by construction \textit{known} and \textit{deterministic}: $\omega(\cdot|z_{t},u_{t})$ is the Dirac delta centered at $z_{t+1}=(u_{1},…,u_{t})$. We have $p(\tau)=\prod_{t}\pi(u_{t}|z_{t})$. However, in imitation learning the dynamics are generally \textit{unknown} and \textit{stochastic}. We have $p(\tau)=\prod_{t}\pi(u_{t}|z_{t})\omega(z_{t}|z_{t-1},u_{t-1})$. This difference is crucial. Consider walking through the logic of Section 3, but applying it to imitation learning instead. Begin with the gradient update for learning the Gibbs parameters (Equation \ref{eq:vebmenergy}). To avoid costly inner-loop optimization of $\theta$ to completion, we again consider (1) importance sampling, as well as our preferred method of (2) contrastive estimation. We will see that in imitation learning neither of these work out of the box. For (1), the importance weights now become $\exp(F_{\phi}(\tau))/p_{\theta}(\tau)$, where $p_{\theta}(\tau)=\prod_{t}\pi_{\theta}(u_{t}|z_{t})\omega(z_{t}|z_{t-1},u_{t-1})$. But the problem is that we do not have access to $\omega$. We can naively \textit{estimate} it beforehand (clearly inadvisable). Or, we can perform a specific \textit{approximation} in the energy model: \smash{$p_{\phi}(\tau)=\exp(F_{\phi}(\tau)-\log Z_{\phi})\approx\tfrac{1}{Z_{\phi,\omega}}\exp(F_{\phi}(\tau))\prod_{t}\omega(z_{t}|z_{t-1},u_{t-1})$} such that $\omega$ cancels out from the expression. Instead of modeling the distribution of trajectories $p(\tau)$ exactly, this now assumes that transition randomness has a limited effect on behavior and that the partition function is constant for all random outcome samples (see e.g. \cite{ziebart2008maximum}). The situation is similar for (2), where it is easy to see that in imitation learning scenarios Equation \ref{eq:classifier} is no longer accessible without performing the above approximation to cancel out $\omega$. Three points bear emphasis. First, if we are willing to make this simplification, we are no longer speaking of TimeGCI anymore. Instead, it is easy to show that this effectively becomes a sort of trajectory-centric ``maximum \textit{causal} entropy'' inverse optimal control (see e.g. \cite{ziebart2010thesis}), which is no longer learning $p(\tau)$ with an exact objective. Second, in general we have little reason to believe that such a trajectory-centric approach would work well in imitation learning: The variance of sample-based estimates is now exacerbated by the unknown and stochastic environment dynamics in imitation learning, such that trajectory-based sampling is generally known to perform quite poorly (see e.g. \cite{fu2018learning}). In fact, for this reason modern adversarial imitation learning methods almost always take a transition-centric approach. Third, empirically a similar structured-classifier approach (with the above approximation) has indeed been found to be unworkable in practice due to high variance (see \cite{fu2018learning} Section 4, implementing \cite{finn2016connection}); moreover, an importance-sampling based approach (with the above approximation) has been made to work, but has required hand-crafted, domain-specific regularization to work, and the learned energy functions only explain the demonstrations locally \cite{finn2016guided}. For these reasons, we do not evaluate TimeGCI on imitation learning scenarios. It is simply not applicable without modifying it to become a different method altogether. Modern imitation learning methods are designed specifically to handle the unknown and stochastic nature of environment transitions. In particular, transition-centric approaches (i.e. scoring $(z,u)$ pairs) have been found to be more effective empirically. What we are doing with TimeGCI, on the other hand, is to focus on an exact objective for learning time-series trajectories $p(\tau)$, which requires a trajectory-centric approach. This is because this allows us to equate TimeGCI with a special kind of noise-contrastive estimation, for which we can enjoy some theoretical guarantees. (Note that these properties are lost when performing the above approximation in an imitation learning setting).}

\subsection{Can AIL be used for Time-series Generation?}\label{app:d2}

\begin{table*}[h!]\small
\renewcommand{\arraystretch}{0.97}
\setlength\tabcolsep{8pt}
\centering
\caption{\textit{Performance Comparison of GAIL, AIRL, and TimeGCI}. Bold numbers indicate best-performing results.}

\vspace{-1.00em}

\label{tab:result2}
\begin{center}
\resizebox{\textwidth}{!}{%
\begin{tabular}{l|l|ccccc}
\toprule
\textit{Benchmark} & \textit{Metric} & Sines & Energy & Gas & Metro & MIMIC-III \\
\midrule
  \multirow{4}{*}{GAIL-SAC}
&Predictive Score&0.447 $\pm$ 0.002&0.261 $\pm$ 0.001&0.022 $\pm$ 0.002&0.257 $\pm$ 0.001&0.017 $\pm$ 0.001\\
&+3 Steps Ahead  &0.472 $\pm$ 0.003&0.262 $\pm$ 0.001&0.048 $\pm$ 0.001&0.250 $\pm$ 0.001&0.012 $\pm$ 0.001\\
&+5 Steps Ahead  &0.542 $\pm$ 0.004&0.261 $\pm$ 0.001&0.074 $\pm$ 0.002&0.245 $\pm$ 0.003&0.011 $\pm$ 0.001\\
&$x$-Corr. Score   &6.798 $\pm$ 0.014&142.457 $\pm$ 0.471&111.124 $\pm$ 0.232&1.044 $\pm$ 0.031&540.89 $\pm$ 0.159\\
\midrule
  \multirow{4}{*}{AIRL-SAC}
&Predictive Score&0.223 $\pm$ 0.006&0.283 $\pm$ 0.002&0.0478 $\pm$ 0.0015&0.243 $\pm$ 0.001&0.018 $\pm$ 0.005\\
&+3 Steps Ahead  &0.381 $\pm$ 0.002&0.295 $\pm$ 0.003&0.0883 $\pm$ 0.0029&0.250 $\pm$ 0.001&0.019 $\pm$ 0.003\\
&+5 Steps Ahead  &0.349 $\pm$ 0.005&0.321 $\pm$ 0.001&0.1176 $\pm$ 0.0040&0.251 $\pm$ 0.002&0.019 $\pm$ 0.001\\
&$x$-Corr. Score   &10.397 $\pm$ 0.005&202.61  $\pm$ 0.119&144.79 $\pm$ 0.3199&1.286 $\pm$ 0.098&2635.57 $\pm$ 0.050\\
\midrule
  \multirow{4}{*}{TimeGCI}
& Predictive Score
& \textbf{0.097} $\pm$ \textbf{0.001} & \textbf{0.251} $\pm$ \textbf{0.001} & \textbf{0.018} $\pm$ \textbf{0.000} & \textbf{0.239} $\pm$ \textbf{0.001} & \textbf{0.002} $\pm$ \textbf{0.000} \\
& +3 Steps Ahead
& \textbf{0.104} $\pm$ \textbf{0.001} & \textbf{0.251} $\pm$ \textbf{0.001} & \textbf{0.042} $\pm$ \textbf{0.001} & \textbf{0.239} $\pm$ \textbf{0.001} & \textbf{0.001} $\pm$ \textbf{0.000} \\
& +5 Steps Ahead
& \textbf{0.109} $\pm$ \textbf{0.001} & \textbf{0.251} $\pm$ \textbf{0.001} & \textbf{0.067} $\pm$ \textbf{0.001} & \textbf{0.239} $\pm$ \textbf{0.001} & \textbf{0.001} $\pm$ \textbf{0.000} \\
& $x$-Corr. Score
& \textbf{1.195} $\pm$ \textbf{0.011} & \textbf{105.2} $\pm$ \textbf{0.433} & \textbf{47.91} $\pm$ \textbf{0.811} & \textbf{0.738} $\pm$ \textbf{0.019} & \textbf{194.3} $\pm$ \textbf{0.180} \\
\bottomrule
\end{tabular}
}
\end{center}

\vspace{-1.00em}

\end{table*}

\dayum{
Conversely to the preceding, we can also ask: Can adversarial imitation learning methods work for time-series generation (hence can we compare them against TimeGCI)? The answer is ``yes'', they can be \textit{applied} to time-series generation, but ``no'', there is no reason to expect they would perform better.
Modern adversarial imitation learning methods come in two broad flavors: (1) GAIL-based \cite{ho2016generative,baram2017model,jeon2018bayesian,zhang2020f}, which do not recover a reward function, and (2) AIRL-based \cite{fu2018learning,qureshi2019adversarial,arenz2020non}, which do recover a reward function. Both are transition-centric (i.e. scoring $(z,u)$ pairs).
For (1): Here, optimization is of the familiar GAN-like objective: $
\text{min}_{\theta}\max_{\phi}\mathbb{E}_{z,u\sim \mu_{s}}\log d_{\phi}(z,u)+\mathbb{E}_{z,u\sim \mu_{\theta}}\log\big(1-d_{\phi}(z,u)\big)
$, which minimizes the JS-divergence between the \textit{state-action} occupancy measures (i.e. distribution of transitions, in time-series generation terms) induced by the learned and source policies. Note that this is the same as the TimeGAN objective \cite{yoon2019time}, modulo entropic/supervised regularization. As noted before, matching the transition marginals is indirectly performing the ``global'' sort of moment-matching. However, as pertains time-series generation, the key difference between GAIL and TimeGAN is that the former performs policy optimization for the generator, whereas the latter simply performs backpropagation. Otherwise, note that both methods are adversarial (viz. saddle-point optimization) and do not learn an explicit energy (viz. black-box discriminator).
In light of this, we may consider GAIL for experimental comparison (to be consistent, we also use SAC as the policy optimization method); see Table \ref{tab:result2} above.
For (2): Here, the discriminator is first constructed to be \textit{transition}-based: $
d_{\theta,\phi}(z,u)=(\exp(g_{\phi}(z,u))/(\exp(g_{\phi}(z,u))+\pi_{\theta}(u|z)))
$. Note that this actually breaks the relationship with modeling the distribution of \textit{trajectories} directly $p_{\phi}(\tau)=\exp(F_{\phi}(\tau)-\log Z_{\phi})$. Importantly, if we apply AIRL to time-series generation, we lose the convergence property that comes from the relationship with noise-contrastive estimation (cf. Proposition \ref{thm:con}), and the gradient of the discriminator is no longer related to the original trajectory-wise energy-based gradient (cf. Proposition \ref{thm:sce}). In fact, all we are left with is that the global optimum is a correct optimum---but this by itself is not helpful (e.g. the global optimum of naive MLE is also a correct optimum, but there is no guarantee that it can be found effectively). In fact, it has been formally shown that the original theoretical justifications for AIRL are incorrect (see e.g. \cite{arenz2020non} Sec. 2.4.2, or \cite{ni2020f} Sec. B.2). (In stochastic domains in imitation learning, it is true that the empirical benefit of variance-reduction from sampling in this transition-based manner still dominates. In time-series generation, however, there is no stochasticity in the ``environment'', and doing it this way may unnecessarily bias our objective of learning $p(\tau)$).
In light of this, we may consider AIRL for experimental comparison (to be consistent, we also use SAC for policy optimization); see Table \ref{tab:result2}.}

\subsection{Ablation Studies on Sequence Lengths}\label{app:d3}

\dayum{
First, we compare the performance of T-Forcing to TimeGCI, using sequence lengths $T\in\{2, 8, 24\}$. We compute the TSTR predictive score, +3 steps ahead, +5 steps ahead, and $x$-Corr. score as usual for each of these settings. Note that for $T=2$ we cannot ask the TSTR evaluation model to predict more than one step ahead, since there is no data for more than one step ahead; we indicate this with ``N/A''. The Gas dataset is used. For ease of interpretation, we consider T-Forcing as the ``baseline'', and also express the TimeGCI numbers as a fraction of the T-Forcing numbers. The results are consistent with what we would expect: The performance advantage enjoyed by TimeGCI over T-Forcing diminishes as the sequence lengths of the input dataset decreases, and increases as the sequence lengths of the input dataset increases. This is true regardless of what metric we are talking about (i.e. 1/3/5-step TSTR score, or feature correlation score). Moreover, at $T=2$ there is almost no difference between the TSTR scores of T-Forcing and TimeGCI. This shows that T-Forcing appears to preserve quite well the one-step ahead relationships of the original data (viz. $T=2$). However, for longer sequences TimeGCI performs relatively better, which is consistent with the motivation behind using an objective that matches the distribution of trajectories, instead of simply matching transition conditionals:}

\begin{table*}[h!]\small
\renewcommand{\arraystretch}{0.97}
\setlength\tabcolsep{8pt}
\centering

\vspace{-0.25em}

\begin{center}
\resizebox{0.7\textwidth}{!}{%
\begin{tabular}{l|l|ccc}
\toprule
\textit{Benchmark} & \textit{Metric} & $T=2$ & $T=8$ & $T=24$ \\
\midrule
  \multirow{4}{*}{T-Forcing}
&Predictive Score&0.038 $\pm$ 0.001&0.036 $\pm$ 0.007&0.035 $\pm$ 0.003\\
&+3 Steps Ahead &N/A&0.076 $\pm$ 0.007&0.080 $\pm$ 0.001\\
&+5 Steps Ahead &N/A&0.114 $\pm$ 0.003&0.111 $\pm$ 0.001\\
&$x$-Corr. Score  &160.3 $\pm$ 0.346&154.2 $\pm$ 0.219&150.8 $\pm$ 0.067\\
\midrule
  \multirow{4}{*}{TimeGCI}
&Predictive Score&0.037 $\pm$ 0.001&0.022 $\pm$ 0.000&0.018 $\pm$ 0.000\\
&+3 Steps Ahead &N/A&0.048 $\pm$ 0.001&0.042 $\pm$ 0.001\\
&+5 Steps Ahead &N/A&0.071 $\pm$ 0.001&0.067 $\pm$ 0.001\\
&x-Corr. Score  &140.4 $\pm$ 0.173&50.72 $\pm$ 0.816&47.91 $\pm$ 0.811\\
\midrule
  \multirow{4}{*}{Ratio of Mean}
&Predictive Score&97.37\%&61.11\%&51.43\%\\
&+3 Steps Ahead &N/A&63.16\%&52.50\%\\
&+5 Steps Ahead &N/A&62.28\%&60.36\%\\
&x-Corr. Score  &87.59\%&32.89\%&31.77\%\\
\bottomrule
\end{tabular}
}
\end{center}

\vspace{-1.25em}

\end{table*}

\dayum{
Second, we use a new synthetic simulation where what is being generated is a set of multi-dimensional sinusoidal waves which---if not perturbed by noise---correspond to different frequencies, phases, etc. This allows us to manually inject noise where we want, and also allows us to simulate ``ground truth''. First, we train both models (T-Forcing and TimeGCI) using the same training data generated by the simulator, and save these learned models. Then during validation, we obtain a sequence from the simulator, then add some additional noise. Specifically, we add independent Gaussian noise $\mathcal{N}(\mu,\sigma)$, using $\mu=0$ and $\sigma=0.1$, to each of the feature dimensions of the wave at time $K$. Then, our task is to predict $t$ steps ahead with the learned models. To be clear, we ask the learned models to predict ahead using the information up to and including this (perturbed) $K$-th step; if the prediction is for $t>1$, we perform open-loop sampling as usual.
We can measure the model's prediction error (i.e. evaluating how well the model forecasts the future based on the provided histories as input, along with the perturbation). Note that this is entirely different from the TSTR predictive score hitherto used (i.e. evaluating how well the model's freely generated output dataset preserves the characteristics of the input dataset); this is only done as an additional sensitivity analysis: By analogy to imitation learning, we can interpret this experiment as a special case of ``action-matching'' (i.e. based on ground-truth state/history as input, we can measure the ``prediction error'' between the action chosen by the imitator policy and the actual action taken by the expert)---but specifically where the state/history has been perturbed with additional noise.
We ask each model to perform $t$-steps ahead prediction, where $t\in\{1,2,3,4,5\}$ as a sensitivity. When we  add noise, we use values $\{\sigma,2\sigma,3\sigma,4\sigma,5\sigma\}$ as a sensitivity. Given any sequence, the step $K$ at which noise is artificially added (and at which point the models are asked to perform $t$-step ahead predictions) is uniformly randomly sampled from $K\in\{1,2,...,T-1\}$.}

\begin{table*}[h!]\small
\renewcommand{\arraystretch}{0.97}
\setlength\tabcolsep{8pt}
\centering

\vspace{-0.25em}

\begin{center}
\resizebox{0.9\textwidth}{!}{%
\begin{tabular}{c|l|ccccc}
\toprule
\textit{Noise at $K$} & \textit{Benchmark} & 1-Ahead MSE & 2-Ahead MSE & 3-Ahead MSE & 4-Ahead MSE & 5-Ahead MSE \\
\midrule
  \multirow{2}{*}{$\sigma$}
&T-Forcing & 0.024 $\pm$ 0.000 & 0.029 $\pm$ 0.001 & 0.036 $\pm$ 0.001 & 0.041 $\pm$ 0.001 & 0.047 $\pm$ 0.001 \\
&TimeGCI   & 0.024 $\pm$ 0.000 & 0.025 $\pm$ 0.001 & 0.026 $\pm$ 0.001 & 0.028 $\pm$ 0.001 & 0.031 $\pm$ 0.001 \\
\midrule
  \multirow{2}{*}{2$\sigma$}
&T-Forcing & 0.055 $\pm$ 0.000 & 0.059 $\pm$ 0.001 & 0.065 $\pm$ 0.001 & 0.071 $\pm$ 0.002 & 0.076 $\pm$ 0.001 \\
&TimeGCI   & 0.055 $\pm$ 0.001 & 0.055 $\pm$ 0.001 & 0.057 $\pm$ 0.001 & 0.059 $\pm$ 0.000 & 0.061 $\pm$ 0.001 \\
\midrule
  \multirow{2}{*}{3$\sigma$}
&T-Forcing & 0.106 $\pm$ 0.001 & 0.109 $\pm$ 0.002 & 0.116 $\pm$ 0.001 & 0.121 $\pm$ 0.001 & 0.127 $\pm$ 0.002 \\
&TimeGCI   & 0.106 $\pm$ 0.001 & 0.106 $\pm$ 0.001 & 0.107 $\pm$ 0.001 & 0.109 $\pm$ 0.001 & 0.111 $\pm$ 0.001 \\
\midrule
  \multirow{2}{*}{4$\sigma$}
&T-Forcing & 0.176 $\pm$ 0.001 & 0.181 $\pm$ 0.002 & 0.185 $\pm$ 0.001 & 0.191 $\pm$ 0.002 & 0.197 $\pm$ 0.002 \\
&TimeGCI   & 0.176 $\pm$ 0.002 & 0.176 $\pm$ 0.002 & 0.177 $\pm$ 0.001 & 0.179 $\pm$ 0.002 & 0.181 $\pm$ 0.001 \\
\midrule
  \multirow{2}{*}{5$\sigma$}
&T-Forcing & 0.266 $\pm$ 0.003 & 0.271 $\pm$ 0.002 & 0.277 $\pm$ 0.002 & 0.280 $\pm$ 0.002 & 0.287 $\pm$ 0.002 \\
&TimeGCI   & 0.266 $\pm$ 0.002 & 0.267 $\pm$ 0.003 & 0.267 $\pm$ 0.002 & 0.270 $\pm$ 0.003 & 0.272 $\pm$ 0.003 \\
\bottomrule
\end{tabular}
}
\end{center}

\vspace{-1.00em}

\end{table*}

\dayum{
The results give some orthogonal intuition as to why the proposed method is better; specifically, is it because of better first-step prediction, or is it because of better robustness when having small errors in the previous steps? Observe that for \textit{single}-step prediction, there is virtually no difference between the performance of T-Forcing and TimeGCI when evaluating the prediction MSE. The more noise added, the worse both models perform; but there is little difference between them, no matter the noise. On the other hand, for \textit{multi}-step prediction, when predicting multiple steps ahead using open-loop sampling, T-Forcing performs worse than TimeGCI. In fact, the gap between their performances increases as $t$ increases. So it appears that it is not the case that TimeGCI simply has better first-step prediction per se; also, it appears that TimeGCI does have better robustness when having errors in the previous steps. Because, while the 1-step performance of both T-Forcing and TimeGCI are impacted almost equally, TimeGCI appears to have an advantage in terms of ``propagating'' less of that error into later time steps.}

\clearpage
\bibliographystyle{unsrt}
\bibliography{
bib/0-imitate,
bib/1-reinforce,
bib/2-entropy,
bib/3-information,
bib/4-constraints,
bib/5-risk,
bib/6-intrinsic,
bib/7-bounded,
bib/8-identification,
bib/9-interpret,
bib/a-miscellaneous,
bib/b-time,
bib/c-yoon
}

\end{document}